\documentclass[10pt,twocolumn,letterpaper]{article}

\usepackage[pagenumbers]{wacv} %

\newcommand{\pluseq}{\mathrel{+}=}

\usepackage[linesnumbered, ruled, vlined]{algorithm2e}
\SetKwInput{KwData}{In}
\SetKwInput{KwResult}{Out}
\SetArgSty{} 
\SetAlCapNameFnt{\footnotesize}
\SetAlCapFnt{\footnotesize}

\usepackage{pgfplots}
\usepackage{pgfplotstable}
\pgfplotsset{compat=newest}
\usepgfplotslibrary{groupplots}
\usepgfplotslibrary{fillbetween}

\usepackage{fontawesome5}

\usetikzlibrary{
    arrows.meta,
    positioning,
    shadows,
    fit,
    calc,
    decorations.pathreplacing,
}

\pgfplotsset{
every axis/.append style={
    title style={font=\footnotesize},
    label style={font=\footnotesize},
    tick label style={font=\footnotesize},
    legend style={font=\footnotesize}    
},
nodes near coords black white/.style={
    small value/.style={
        font=\footnotesize,
        anchor=center,
        text=white,
        xshift=-0.05cm,        
    },
    large value/.style={
        font=\footnotesize,
        anchor=center,
        text=black,
        xshift=-0.05cm,
    },
    every node near coord/.style={
        check for zero/.code={
            \pgfmathfloatifflags{\pgfplotspointmeta}{0}{
            \pgfkeys{/tikz/coordinate}
            }{
            \begingroup
            \pgfkeys{/pgf/fpu}
            \pgfmathparse{\pgfplotspointmetatransformed<700}
            \global\let\result=\pgfmathresult
            \endgroup
            \pgfmathfloatcreate{1}{1.0}{0}
            \let\ONE=\pgfmathresult
            \ifx\result\ONE
                \pgfkeysalso{/pgfplots/small value}
            \else
                \pgfkeysalso{/pgfplots/large value}
            \fi
            }
        },
        check for zero,
    },
},
hdist style/.style={
    every tick label/.append style={font=\normalsize},
    nodes near coords={
        \begingroup
        \pgfkeys{/pgf/fpu}
        \pgfmathparse{\pgfplotspointmeta > 10}
        \global\let\result=\pgfmathresult
        \endgroup
        
        \pgfmathfloatifflags{\result}{1}{
            \pgfmathprintnumber[fixed, fixed zerofill, precision=1]{\pgfplotspointmeta}
        }{
           \pgfmathprintnumber[fixed, fixed zerofill, precision=2]{\pgfplotspointmeta} 
        }

    },
    nodes near coords black white,
},
hdist style integers/.style={
    every tick label/.append style={font=\normalsize},
    nodes near coords={
        \begingroup
        \pgfkeys{/pgf/fpu}
        \pgfmathparse{
            \pgfplotspointmeta > 10000 ? round(\pgfplotspointmeta / 1000) : (\pgfplotspointmeta > 1000 ? \pgfplotspointmeta / 1000 : \pgfplotspointmeta)
        }
        \global\let\result=\pgfmathresult
        \pgfmathparse{\pgfplotspointmeta > 10000}
        \global\let\nodegttenk=\pgfmathresult
        \pgfmathparse{\pgfplotspointmeta > 1000}
        \global\let\nodegtk=\pgfmathresult        
        \endgroup

        \pgfmathfloatifflags{\nodegttenk}{1}{
            \pgfmathprintnumber{\result}k%
        }{
            \pgfmathfloatifflags{\nodegtk}{1}{
                \pgfmathprintnumber[fixed, precision=1]{\result}k%
            }{
                \pgfmathprintnumber{\result}%
            }
        }
    },    
    nodes near coords black white,
},
hdist base/.style={
    colormap name=viridis,
    enlargelimits=false,
    title style={yshift=0cm},
    label style={font=\footnotesize},
    xticklabel style={yshift=-4pt},
    yticklabel style={xshift=-4pt},        
    every tick label/.append style={font=\footnotesize},    
},
empty colorbar/.style={
    colorbar,  
    colorbar style={
        nodes near coords={},
        ytick=\empty,
   }
}
}

\usepackage{multirow}

\usepackage{arydshln}

\makeatletter
\def\adl@drawiv#1#2#3{%
        \hskip.5\tabcolsep
        \xleaders#3{#2.5\@tempdimb #1{1}#2.5\@tempdimb}%
                #2\z@ plus1fil minus1fil\relax
        \hskip.5\tabcolsep}
\newcommand{\cdashlinelr}[1]{%
  \noalign{\vskip\aboverulesep
           \global\let\@dashdrawstore\adl@draw
           \global\let\adl@draw\adl@drawiv}
  \cdashline{#1}
  \noalign{\global\let\adl@draw\@dashdrawstore
           \vskip\belowrulesep}}
\makeatother

\usepackage{xcolor}

\definecolor{myteal}{RGB}{0, 128, 128}
\definecolor{myorange}{RGB}{230, 120, 20}
\definecolor{myblue}{RGB}{70, 130, 180}
\definecolor{mygreen}{RGB}{46, 139, 87}
\definecolor{mypurple}{RGB}{148, 0, 211}
\definecolor{mygray}{RGB}{120, 120, 120}

\definecolor{myred}{RGB}{200, 30, 30}
\definecolor{mydarkred}{RGB}{139, 0, 0}
\definecolor{mygold}{RGB}{218, 165, 32}
\definecolor{myamber}{RGB}{255, 191, 0}

\definecolor{mydarkblue}{RGB}{25, 25, 112}
\definecolor{mynavy}{RGB}{0, 0, 128}
\definecolor{myskyblue}{RGB}{135, 206, 235}
\definecolor{mycyan}{RGB}{0, 139, 139}

\definecolor{mylime}{RGB}{50, 205, 50}
\definecolor{mydarkgreen}{RGB}{0, 100, 0}
\definecolor{myolive}{RGB}{128, 128, 0}

\definecolor{myviolet}{RGB}{138, 43, 226}
\definecolor{myfuchsia}{RGB}{255, 0, 255}
\definecolor{myrose}{RGB}{199, 21, 133}

\definecolor{mybrown}{RGB}{139, 69, 19}
\definecolor{mytan}{RGB}{210, 180, 140}
\definecolor{mysilver}{RGB}{192, 192, 192}
\definecolor{myblack}{RGB}{30, 30, 30}

\colorlet{lcolor}{mygreen}
\colorlet{ulcolor}{myorange}

\definecolor{wacvblue}{rgb}{0.21,0.49,0.74}
\usepackage[pagebackref,breaklinks,colorlinks,allcolors=wacvblue]{hyperref}

\def\wacvPaperID{2176} %
\def\confName{WACV}
\def\confYear{2026}

\title{Semi-Supervised Hierarchical Open-Set Classification}

\author{Erik Wallin\textsuperscript{1,2}, Fredrik Kahl\textsuperscript{2}, Lars Hammarstrand\textsuperscript{2} \\
\textsuperscript{1}Saab AB, \textsuperscript{2}Chalmers University of Technology \\
{\tt\small \{walline,fredrik.kahl,lars.hammarstrand\}@chalmers.se}}

\begin{document}
\maketitle
\begin{abstract}

Hierarchical open-set classification handles previously unseen classes by assigning them to the most appropriate high-level category in a class taxonomy. We extend this paradigm to the semi-supervised setting, enabling the use of large-scale, uncurated datasets containing a mixture of known and unknown classes to improve the hierarchical open-set performance. To this end, we propose a teacher-student framework based on pseudo-labeling. Two key components are introduced: 1) subtree pseudo-labels, which provide reliable supervision in the presence of unknown data, and 2) age-gating, a mechanism that mitigates overconfidence in pseudo-labels. Experiments show that our framework outperforms self-supervised pretraining followed by supervised adaptation, and even matches the fully supervised counterpart when using only 20 labeled samples per class on the iNaturalist19 benchmark. Our code is available at \url{https://github.com/walline/semihoc}.

\end{abstract}
    
\section{Introduction}
\label{sec:intro}

Out-of-distribution (OOD) detection is important for deploying deep learning in real-world settings. Most prior work studies the \emph{binary} case, where a model is trained on a set of in-distribution (ID) classes, and the test-time task is to detect any sample not belonging to these classes \cite{hendrycks2016baseline, liang2017enhancing, sun2022out}.

More recently, researchers have explored the hierarchical setting of OOD detection \cite{lee2018hierarchical, pyakurellearning, wallin2025prohoc}, in which the ID classes are organized according to a known taxonomy or tree structure, and OOD samples that belong somewhere in this hierarchy (but not in the leaf ID classes) should be classified as the correct internal category. This setting, sometimes referred to as hierarchical novelty detection (HND), provides more informative OOD predictions than binary rejection. However, the problem is considerably more challenging, as it effectively becomes a multi-class classification problem over internal categories for which no labeled training data are available.

\begin{figure}[!t]
    \centering
    \input{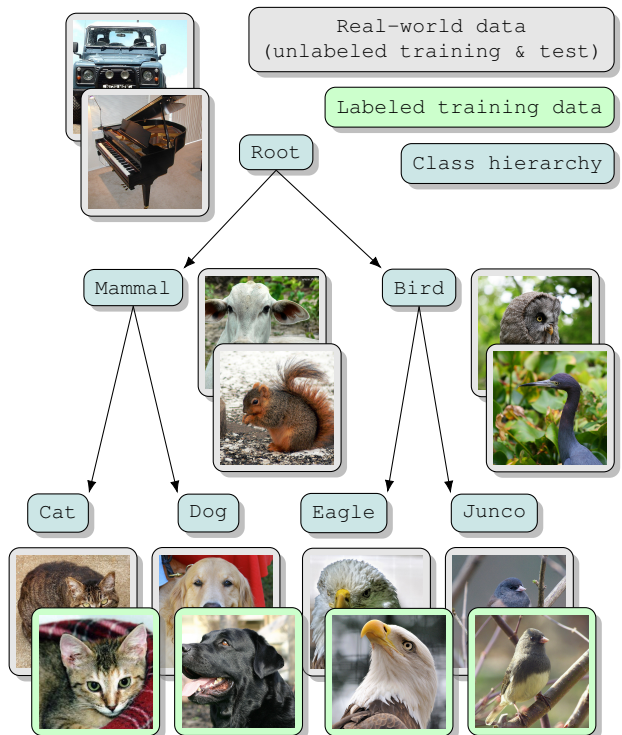} \vspace*{-0.2cm}
    \caption{Semi-supervised hierarchical open-set classification. The goal is to predict data from the open-world distribution (gray) to the appropriate hierarchy nodes.} \vspace*{-0.5cm}
    \label{fig:introfig}
\end{figure}

In parallel, works in domains such as open-set semi-supervised learning \cite{fan2023ssb, wallin2024prosub} and semantically coherent OOD detection \cite{yang2021semantically, lu2023uncertainty} have shown that OOD detection can benefit from exposing the model to OOD samples in the unlabeled training data. This raises the question: if real-world data inherently contain a mix of both ID and OOD samples, why not leverage these data to improve performance? Motivated by this, we introduce \emph{semi-supervised hierarchical open-set classification}, a new problem setting that combines HND with the semi-supervised paradigm (see \cref{fig:introfig}). We aim to explore both the opportunities and the challenges that arise in this setting.

From our investigations, we found that a key challenge is that standard confidence-based pseudo-labeling is unreliable when assigning OOD data to high-level categories, as OOD samples are difficult to distinguish from uncertain ID samples. To address this, we introduce \emph{subtree pseudo-labels} which, when assigned, indicate the likely membership within a specific subtree, without committing to a specific class, be it ID or OOD. This approach provides a more robust supervision signal for OOD samples.

Furthermore, we observe that OOD samples tend to receive increasingly deep pseudo-labels as training progresses, eventually overshooting the correct internal node and confusing it with some overly specific category. To counter these overpredictions, we propose \emph{age-gating}: a mechanism that prevents pseudo-labels that are assigned unexpectedly late during training.

We integrate subtree pseudo-labels and age-gating into a teacher-student framework for semi-supervised hierarchical open-set classification, which we call \emph{SemiHOC}. Experiments demonstrate that SemiHOC outperforms the baseline of self-supervised pretraining followed by supervised adaptation. Notably, on the iNaturalist19 benchmark, SemiHOC matches the performance of full ID supervision while using only 20 labels per class.

Our main contributions are as follows:
\begin{itemize}
    \item We introduce semi-supervised hierarchical open-set classification, which extends HND to the semi-supervised setting, where the unlabeled data contain both ID and OOD samples.
    \item We identify two key challenges of this setting: 1) reliable pseudo-labeling of OOD data and 2) accumulation of overconfidence for OOD data, and address these with our \emph{subtree pseudo-labels} and \emph{age-gating}.
    \item We present SemiHOC, a teacher-student framework for semi-supervised open-set classification that integrates subtree pseudo-labels and age-gating, and show that it can match a fully supervised baseline with only a small number of labels per class.
\end{itemize}

\section{Related work}
\label{sec:related-work}

\subsection{OOD detection}

OOD detection is a well-studied problem in deep learning. In its most common form, a model is trained on a set of in-distribution (ID) classes; during deployment, the goal is to detect inputs that do not belong to any of these classes \cite{hendrycks2016baseline, liang2017enhancing, li2020energy, yang2024generalized}. The standard paradigm addresses this in a binary fashion: a sample is either ID or OOD, disregarding semantic similarities between the OOD data and the ID classes. In contrast, the hierarchical settings of HND and this work allow for more informative predictions by assigning OOD samples to the most appropriate internal node of a given class hierarchy.

\subsection{Class hierarchies for classification}

Many classification problems involve classes that can be organized into a hierarchy, such as biological taxonomy or product categorization. A common approach is to train models that optimize hierarchical metrics, such as tree distances or hierarchical precision \cite{bertinetto2020making,karthik2021no,jain2024test,kosmopoulos2015evaluation,garg2022learning,liang2023inducing}. These methods account for the severity of classification errors within the hierarchy but assume that all test samples belong to known classes, and thus do not address OOD data.

\subsection{OOD detection with class hierarchies}

Several works have addressed the classification of OOD samples as internal nodes in a class hierarchy \cite{lee2018hierarchical,ruiz2022hierarchical,pyakurel2024hierarchical,pyakurellearning,wallin2025prohoc,linderman2023fine}. This setting has been referred to as hierarchical novelty detection \cite{lee2018hierarchical}, hierarchical OOD detection \cite{wallin2025prohoc}, and fine-grained openset detection \cite{pyakurellearning}. We use the term \emph{hierarchical open-set classification}, which better fits the semi-supervised setting where the unknown classes appear in the unlabeled data (\ie, they are not fully novel), and highlights that the goal is classification of unknown classes rather than binary rejection. There are two baseline approaches for this problem: top-down and flattening.

In the top-down approach \cite{lee2018hierarchical,linderman2023fine}, local classifiers and OOD detectors are placed at each internal node in the hierarchy. At inference time, the model traverses from the root, predicting at each node whether the sample is locally OOD based on a thresholded score. The sample is assigned to the deepest node before exceeding the threshold. While intuitive, this approach has the drawback of requiring tuned thresholds for each OOD detector, which is impractical in real-world scenarios without access to labeled OOD data.

The flattening approach \cite{lee2018hierarchical, ruiz2022hierarchical, pyakurel2024hierarchical} instead trains a single flat classifier to directly predict among both ID leaf classes and internal OOD nodes. Since no OOD training data are available for the internal nodes, the usual workaround is to use data from descendant ID classes as a proxy. However, this may poorly represent the true distribution of OOD samples at internal nodes.

More recently, methods that more explicitly leverage the hierarchical structure have been proposed in ProHOC \cite{wallin2025prohoc} and State-FGOD \cite{pyakurellearning}. ProHOC utilizes the class hierarchy to factorize the distribution of predictions, such that each prediction becomes a product of conditional probabilities corresponding to the path from the root to the predicted node. The conditional distributions are modeled from ID classification networks trained at different levels of the hierarchy. State-FGOD proposes a state-based representation in which a network is trained to predict logits for each node in the hierarchy. The logits corresponding to valid states (paths in the class hierarchy) are multiplied and transformed to a predictive distribution using the softmax function.

A drawback of many existing methods for hierarchical novelty detection is their reliance on bias parameters to set the trade-off between ID and OOD performance. These biases are difficult to tune in practice without access to labeled OOD data. In this work, we therefore build on the ProHOC methodology, which is designed to avoid such tuning.

\subsection{Unlabeled ID/OOD exposure}

OOD detection is motivated by the fact that real-world data often contains samples outside the target classes. Yet most traditional approaches assume access only to ID data during training, treating OOD detection purely as a test-time task. If, however, real-world data contain both ID and OOD, it is natural to exploit such unlabeled and uncurated data to improve both ID classification and OOD detection.

Recent research directions address this more realistic setting by leveraging an unlabeled mix of ID and OOD data during training. Examples include semantically coherent OOD (SC-OOD) detection \cite{yang2021semantically, lu2023uncertainty}, open-set semi-supervised learning (OsSSL) \cite{li2023iomatch, fan2023ssb, wallin2024prosub, ma2023rethinking}, and open-world semi-supervised learning (OwSSL) \cite{cao2022open, rizve2022towards, rizve2022openldn, liu2023open}.

SC-OOD detection highlights that many benchmarks use distinct datasets for ID and OOD, disregarding any semantic overlap between these datasets. For example, dogs from Tiny ImageNet are treated as OOD when the model is trained on CIFAR-10, which also contains dogs. SC-OOD trains on an unlabeled mix of both datasets, enabling the model to recognize overlapping categories as ID.

OsSSL assumes a small labeled ID set and a larger unlabeled mix of ID and OOD data, leveraging the unlabeled data to improve both ID classification and OOD detection. OwSSL has a similar data setup to OsSSL but differs in goal: OwSSL aims not only to detect OOD samples, but also to classify them into distinct classes. However, OwSSL typically relies on strong assumptions, such as balanced OOD class distributions or a known number of OOD classes, which limit its practicality.

Our work also utilizes an unlabeled ID/OOD mix but differs from existing approaches by:
\begin{itemize}
    \item incorporating relationships between classes through a class hierarchy,
    \item performing OOD classification as opposed to binary detection, predicting OOD as internal categories of the hierarchy, and
    \item making no assumptions about the number or distribution of OOD classes, some categories may have abundant OOD data, others none.
\end{itemize}

\section{Semi-supervised hierarchical\texorpdfstring{\\}{ }open-set classification}
\label{sec:method}

\subsection{Problem formulation}
\label{sec:prelim}

We consider a labeled training set with samples from domain $\mathcal{X}$ with classes $\mathcal{C}^\text{id}$. These classes form the leaf nodes of a known directed rooted tree (hierarchy) $\mathcal{H}$ with node set $\mathcal{C}$ such that $\mathcal{C}^\text{id} \subset \mathcal{C}$. The labeled set is drawn from the ID distribution $p^\text{id}(x,y)$ where $x \in \mathcal{X}$ is the data and $y \in \mathcal{C}^\text{id}$ is the label. The unlabeled training set and the test set are drawn from an open-world distribution $p^\text{open}(x,y)$ with $y \in C$ for which we only observe $x \in \mathcal{X}$. Thus, $p^\text{open}$ covers both ID samples ($y \in \mathcal{C}^\text{id}$) and OOD samples ($y \in \mathcal{C} \setminus \mathcal{C}^\text{id}$), where OOD labels correspond to internal nodes of $\mathcal{H}$. Our objective is to learn a classifier, $f: \mathcal{X} \rightarrow \mathcal{C}$, that predicts the correct node in $\mathcal{H}$: a leaf if the sample belongs to an ID class, or the appropriate internal node if the sample is OOD.

\subsection{ProHOC} \label{sec:prohoc}

This work builds on ProHOC \cite{wallin2025prohoc}, a method for HND trained with labeled data from $p^\text{id}$ and evaluated on the open-set distribution $p^\text{open}$. ProHOC uses depth-specific classification networks, one per hierarchy depth, each trained with standard cross-entropy using ID data whose labels are remapped to their ancestors at the specified depth. The depth-specific predictions are fused into a hierarchical distribution by modeling conditional probabilities at each internal node, based on the confidences and entropies of the depth-specific predictions. This yields a probability distribution over the union of all possible OOD predictions (internal nodes) and ID classes (leaf nodes). In this work, we use ProHOC's hierarchical predictions for pseudo-labeling, and we use these pseudo-labels to train the depth-specific classification models with standard cross-entropy losses.

\subsection{Pseudo-labeling}
\label{sec:pseudo-labeling}

A core technique for semi-supervised learning is pseudo-labeling \cite{lee2013pseudo}: using unlabeled samples with high-confidence predictions as training data. This requires a correlation between the prediction confidence and the accuracy, allowing us to assign pseudo-labels with high precision. This correlation is established for the ID case \cite{sohn2020fixmatch}, but does not necessarily need to hold for OOD predictions in hierarchical open-set classification.

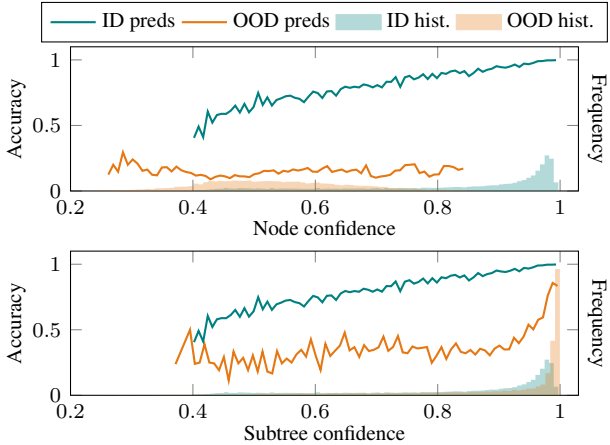
\begin{figure}
    \centering
    \pgfplotsset{
    suppress low frequency/.style args={#1}{
        x filter/.expression={
            \thisrow{frequency} < #1 ? NaN : x
            }
    }    
}

\begin{tikzpicture}
\begin{groupplot}[
  group style={
    group size=1 by 2,
    vertical sep=0.8cm,
  },
  width=\columnwidth,
  height=3.5cm,
  xmin=0.2, xmax=1.03,
  ymin=0, ymax=1.1,
  clip=true,
  clip mode=individual,
  ticklabel style = {font=\footnotesize},
  label style={font=\footnotesize},
  ylabel style={yshift=-0.15cm},
  legend columns=-1,
  legend style={at={(0.5,1.03)}, anchor=south, font=\footnotesize},
]

\nextgroupplot[
  title={},
  ylabel={Accuracy},
  xlabel={Node confidence},
  xlabel style={yshift=0.15cm},
]
\addplot+[mark=none, thick, color=myteal] table[
    x=bin_center,
    y=accuracy,
    suppress low frequency={0.001},
    col sep=comma,
] {data/id_confs_epoch299.csv};
\addlegendentry{ID preds}

\addplot+[mark=none, thick, color=myorange] table[
    x=bin_center,
    y=accuracy,
    suppress low frequency={0.001},    
    col sep=comma,
] {data/ood_confs_epoch299.csv};
\addlegendentry{OOD preds}

\addplot [name path=ZERO,draw=none, forget plot] coordinates {(0,0) (1,0)};

\addplot[color=myteal, draw=none, name path=ID, y filter/.expression={y * 3}, forget plot]
table[col sep=comma, x=x, y=y] {data/id_confs_epoch299_stairs.csv};
\addplot [myteal, opacity=0.3] fill between [of=ID and ZERO];
\addlegendentry{ID hist.}

\addplot[color=myorange, draw=none, name path=OOD, y filter/.expression={y * 3}, forget plot]
table[col sep=comma, x=x, y=y] {data/ood_confs_epoch299_stairs.csv};
\addplot [myorange, opacity=0.3] fill between [of=OOD and ZERO];
\addlegendentry{OOD hist.}

\node[rotate=-90, anchor=south, color=black,] at (axis description cs: 1.00, 0.5) {\footnotesize Frequency};

\nextgroupplot[
  title={},
  xlabel={Subtree confidence},
  ylabel={Accuracy},
  xlabel style={yshift=0.15cm},  
]
\addplot+[mark=none, thick, color=myteal] table[
    x=bin_center,
    y=accuracy,
    suppress low frequency={0.001},     
    col sep=comma,
] {data/id_cumconfs_epoch299.csv};

\addplot+[mark=none, thick, color=myorange] table[
    x=bin_center,
    y=accuracy,
    suppress low frequency={0.001},     
    col sep=comma,
] {data/ood_cumconfs_epoch299.csv};

\addplot [name path=ZERO,draw=none, forget plot] coordinates {(0,0) (1,0)};

\addplot[color=myteal, draw=none, name path=ID, y filter/.expression={y * 3}, forget plot]
table[col sep=comma, x=x, y=y] {data/id_cumconfs_epoch299_stairs.csv};
\addplot [myteal, opacity=0.3] fill between [of=ID and ZERO];

\addplot[color=myorange, draw=none, name path=OOD, y filter/.expression={y * 3}, forget plot]
table[col sep=comma, x=x, y=y] {data/ood_cumconfs_epoch299_stairs.csv};
\addplot [myorange, opacity=0.3] fill between [of=OOD and ZERO];

\node[rotate=-90, anchor=south, color=black,] at (axis description cs: 1.00, 0.5) {\footnotesize Frequency};

\end{groupplot}
\end{tikzpicture} \vspace*{-0.5cm}
    \caption{Average accuracy as a function of prediction confidence for samples predicted as ID and OOD. The shaded regions show the corresponding confidence histograms. The subtree confidence enables pseudo-labeling for OOD-predictions by achieving high accuracy for high-confidence predictions.} \vspace*{-0.5cm}
    \label{fig:conf-vs-acc}
\end{figure}

In the top panel of \cref{fig:conf-vs-acc}, we look at the accuracies as a function of confidence for predictions from a fully trained ProHOC model on iNaturalist19, where ID predictions (leaf nodes) and OOD predictions (internal nodes) are separated. For data predicted as an ID class, we see a clear correlation between prediction confidence and accuracy, indicating that these predictions can be used for pseudo-labeling. However, for data predicted as an internal node, there is no such correlation, making the naive application of confidence-based pseudo-labels unfit for these predictions.

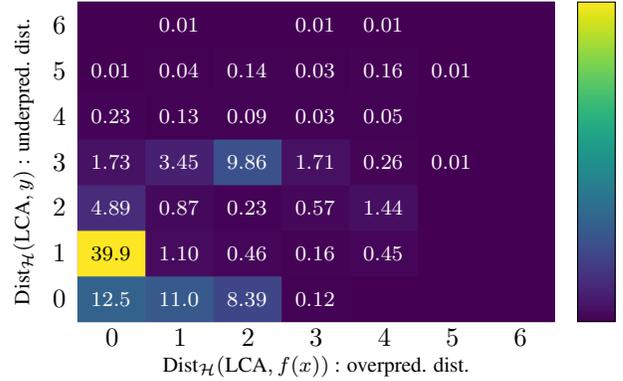
\begin{figure}
    \centering
    \begin{tikzpicture}

\begin{axis}[
    width=0.95\columnwidth,
    height=0.7\columnwidth,
    ymin=-0.5,
    ymax=6.5,
    xmin=-0.5,
    xmax=6.5,    
    xtick={0,1,2,3,4,5,6},
    ytick={0,1,2,3,4,5,6},
    xlabel={$\text{Dist}_\mathcal{H}(\text{LCA}, f(x))$ : overpred. dist.},
    ylabel={$\text{Dist}_\mathcal{H}(\text{LCA}, y)$ : underpred. dist.},
    hdist base,
    hdist style,
    yticklabel style={xshift=3pt},
    xticklabel style={yshift=5pt},    
    xlabel style={yshift=3pt},
    empty colorbar,
    title={},
]

\addplot [
    matrix plot*,
    mesh/cols=7,
    point meta=explicit,
] table [meta=value, col sep=comma] {data/ood_preds_hdists_epoch299.csv};

\end{axis}

\end{tikzpicture} \vspace*{-0.5cm}
    \caption{Distances (the number of edges in $\mathcal{H}$) to the ground truth for data predicted as OOD (internal nodes), displayed as percentages. LCA denotes the Lowest Common Ancestor of the prediction and the ground-truth.} \vspace*{-0.5cm}
    \label{fig:oodpreds-dists}
\end{figure}

However, as shown in \cref{fig:oodpreds-dists}, when analyzing mistakes for samples predicted as OOD, specifically by looking at the distances from the predictions to the ground truth, decomposed into overprediction and underprediction, we see that 59.2\% of samples have an overprediction distance of zero, meaning these are predicted as the correct node or an ancestor of the correct node. The most common being predicting the parent of the ground truth node, which occurs in 39.9\% of OOD predictions.

Motivated by these findings, we propose a new type of pseudo-labels for hierarchical open-set classification: \emph{subtree pseudo-labels} (SPL). A subtree pseudo-label for a node in the hierarchy says that the sample likely belongs to the subtree starting from this node, \ie, the sample is either OOD at this node or one of the descendant nodes, or belongs to one of the descendant leaves. Formally, we define the subtree as
\begin{equation} \label{eq:subtree}
    \text{Subtree}(c) = \{c' \in \mathcal{C} ~|~c'~\text{is a descendant of}~c~\text{or is}~c\},
\end{equation}
where $\mathcal{C}$ is the set of nodes in the class hierarchy $\mathcal{H}$.

The confidence used for SPLs is calculated by summing the predicted probabilities, $p(c|x)$, belonging to the subtree. As shown in the bottom panel of \cref{fig:conf-vs-acc}, subtree confidence enables high accuracy at high confidence for OOD (non-leaf) predictions, allowing us to assign these less specific but more robust SPLs with high precision.

Samples get the subtree pseudo-label if the confidence exceeds a threshold, $\tau$. The subtree pseudo-labels can be formulated as
\begin{equation} \label{eq:subtree-pl}
    \hat{y}_c = \left[ \sum_{c' \in \text{Subtree}(c)} p(c|x) \right] > \tau,
\end{equation}
such that $\hat{y}_u = 1$ if the condition holds and zero otherwise. We denote the set of subtree pseudo-labels for a sample $\mathbf{\hat{y}}$.  

\subsection{Overpredictions of OOD data} \label{sec:age-gating}

It is known from prior work on OsSSL \cite{fan2023ssb} that applying standard confidence-based pseudo-labeling to unlabeled data containing unseen classes eventually causes the unknown samples to be assigned to their most similar ID classes. For hierarchical open-set classification, this risk is even greater, as the unknown classes can appear deep in the hierarchy and thus be semantically and visually close to the ID classes. Moreover, as observed in \cref{sec:pseudo-labeling}, standard confidence-based pseudo-labeling struggles to reliably separate ID from OOD in the hierarchical setting.

Our experiments confirm this problem. \Cref{fig:purity-vs-depth} of \cref{sec:age-gating} shows the purity and average depth of subtree pseudo-labels assigned to OOD samples when training a standard SSL framework with subtree pseudo-labels. Here, purity is defined as the fraction of samples whose ground-truth label lies within the assigned subtree. We see that the model gradually becomes overconfident, assigning OOD samples to increasingly deep nodes in the hierarchy.

\begin{figure}[t]
    \centering
    \input{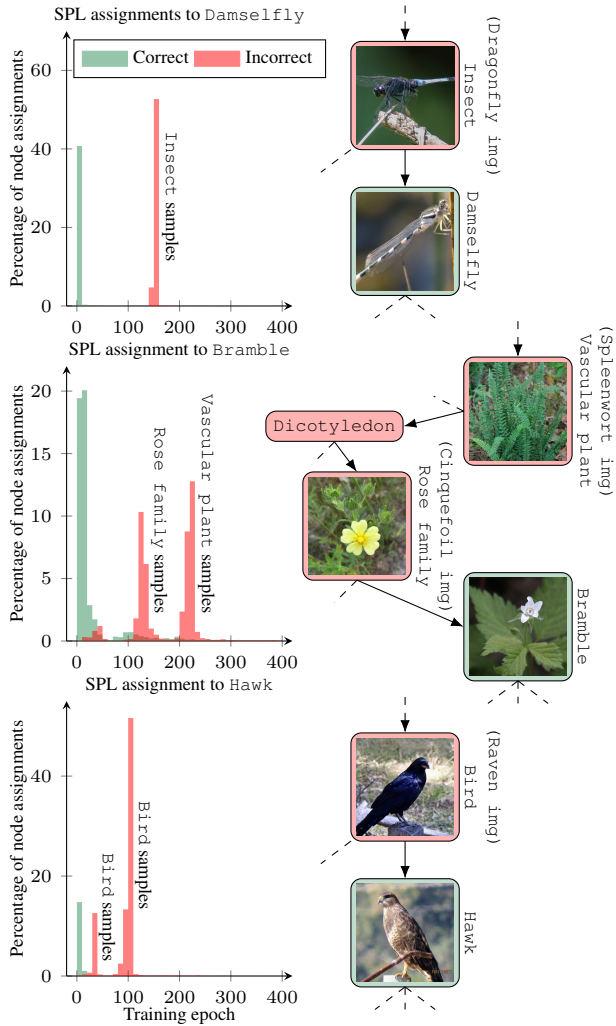} \vspace*{-0.5cm}
    \caption{Frequencies of subtree pseudo-label assignments across training epochs for selected hierarchy nodes, split into correct and incorrect assignments. Each node's local hierarchy structure and example images from correct and incorrect assignments are shown to the right. We see that the majority of correct SPLs are assigned early in training, while incorrect overpredictions of OOD data emerge later as distinct peaks.} \vspace*{-0.5cm}
    \label{fig:assignment-histogram}    
\end{figure}

To further analyze this overconfidence, we study the dynamics of SPL assignments over training epochs in search of informative patterns. For this purpose, we log the first assignment per sample per node and visualize the resulting frequencies, decomposed into correct and incorrect assignments. The dynamics in \cref{fig:assignment-histogram} reveal that correct assignments occur early, while incorrect, overly deep assignments emerge later as distinct peaks. The overpredictions are largely driven by visually similar OOD samples that belong higher in the hierarchy. These results are from iNat19 with 10 labels per class.

This observation motivates our proposed \emph{age-gating} strategy for SPL. The key idea is to prevent additional SPLs from being assigned to a node once the initial wave of correct assignments has passed. We implement this with a simple cutoff-detection method, detailed in \cref{alg:peak-detection}. The cutoff detection scans for the first peak in the SPL assignment frequency of each node and defines a cutoff when the frequency falls below a fixed fraction of the peak value. Cutoff detection is performed per node at the end of each epoch. When a cutoff is set, no new SPLs are assigned to that node. Additional results on the effectiveness of age-gating are presented in \cref{sec:results-age-gating}.

\subsection{Our framework: SemiHOC}

Equipped with our proposed subtree pseudo-labels and age-gating strategy, we introduce \emph{SemiHOC}, a framework for semi-supervised open-set classification. At its core, SemiHOC follows a teacher-student architecture, extended with our components specifically designed for the hierarchical open-set setting. We adopt the ProHOC method to produce hierarchical predictions, as it is the only existing HND method that does not require tuning of an ID/OOD trade-off parameter, a parameter that is difficult to set without access to labeled OOD data.

For the labeled data, we follow the same approach as supervised training for ProHOC: labels are remapped to every height of the hierarchy. These remapped labels are then used with a standard cross-entropy loss to train the corresponding depth-specific networks.

\begin{algorithm}[t]
    \scriptsize
    \caption{Detect cutoffs for age-gating} \label{alg:peak-detection}
    \DontPrintSemicolon
    \SetKwData{Epochs}{epochs}
\SetKwData{Counts}{counts}
\SetKwData{Countsi}{counts$[i]$}
\SetKwData{Edges}{edges}
\SetKwData{Edgesi}{edges$[i]$}
\SetKwData{None}{none}
\SetKwData{MaxCount}{max\_count}

\SetKwFunction{BinCounts}{BinCounts}
\SetKwFunction{Length}{length}

\KwIn{List of pseudo-label assignment epochs for one node \Epochs; bin width $w$; drop threshold $\gamma$}
\KwOut{Step value where a significant drop is detected, or $\infty$}

\BlankLine
\Counts,~\Edges $\leftarrow$ \BinCounts(\Epochs, bin width $w$)

\MaxCount $\leftarrow 0$\;

\For{$i\leftarrow 0$ \KwTo \Length(\Counts)~$-1$}{
    \If{\Countsi $>$ \MaxCount}{
        \MaxCount $\leftarrow$ \Countsi\;
    }
    \ElseIf{\Countsi $< \gamma \cdot$ \MaxCount}{
        \Return \Edgesi
    }
}

\Return $\infty$

\end{algorithm}

For the unlabeled data, the teacher model generates predictions using ProHOC. These predictions are reduced to subtree pseudo-labels following \eqref{eq:subtree-pl}. To support age-gating, we maintain an SPL log as a mapping $(c,g) \rightarrow e$, where $c$ is a class, $g$ is a unique sample identifier, and $e$ is the epoch of assignment. New sample-class pairs are added on the first assignment and removed if not reassigned in subsequent epochs (\ie, if confidence decreases or the prediction switches branch).

We then remove any subtree pseudo-labels assigned \emph{after} the cutoff epoch, as determined by the age-gating procedure. Note that a sample is not necessarily excluded entirely from training. Rather, one or more of its deepest subtree pseudo-labels are blocked and ignored in loss computation.

\begin{figure*}[t]
    \centering
    \input{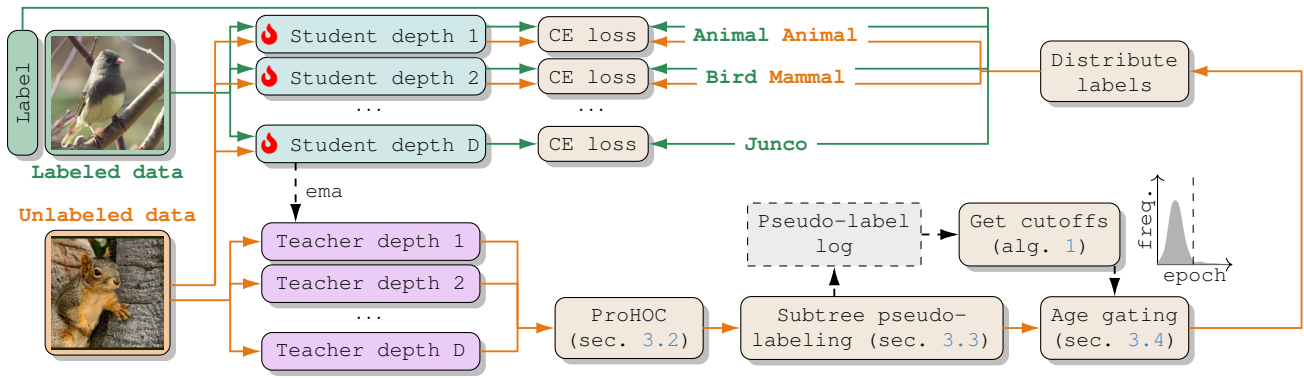}
    \caption{The flow of \textcolor{lcolor}{labeled} and \textcolor{ulcolor}{unlabeled} data in SemiHOC. Labeled data are used to train depth-specific student models with cross-entropy losses using labels mapped to the specified depths. Unlabeled data are predicted by teacher models via ProHOC, producing hierarchical distributions that form subtree pseudo-labels. After age-gating, these subtree pseudo-labels are mapped to the appropriate depths and used to train the students with cross-entropy losses.} \vspace*{-0.5cm}
    \label{fig:semihoc}
\end{figure*}

The remaining subtree pseudo-labels are used to train the student model. If a sample's subtree pseudo-labels extend to depth $d$ in the hierarchy, the corresponding top $d$ networks contribute to its loss via standard cross-entropy, while deeper networks are unaffected.

At the end of each epoch (\ie, a full iteration through the unlabeled data), we apply the cutoff-detection method in \cref{alg:peak-detection} to all hierarchy nodes to determine new cutoff epochs. The flow of data in SemiHOC is visualized in \cref{fig:semihoc} and a detailed training algorithm is provided in the supplementary material.

We use DinoV2 features \cite{oquab2023dinov2}, a widely used image backbone, as input to our depth-specific models. DinoV2 is trained fully self-supervised, ensuring that our approach does not exploit any label information beyond the small labeled training set. The teacher model is obtained by taking an exponential moving average of the student model parameters. We also apply dropout \cite{srivastava2014dropout} to the input features and the intermediate features of the student model, analogous to how strong data augmentations are applied to student inputs in, \eg, the seminal SSL method FixMatch \cite{sohn2020fixmatch}.

\section{Experiments and results} \label{sec:experiments}

\subsection{Datasets}

We evaluate our method on three large-scale image classification benchmarks with class hierarchies: iNaturalist19 \cite{van2018inaturalist}, iNaturalist21-Aves \cite{inaturalist21}, and SimpleHierImageNet \cite{wallin2025prohoc}, the latter constructed from ImageNet \cite{russakovsky2015imagenet} and the WordNet graph \cite{miller1995wordnet}. The full iNaturalist21 dataset contains 10K classes and 2.7M images. To make it more accessible for experimentation, we use the subtree starting from \emph{Class: Aves}, \ie, all bird classes. This bird subset contains 1.5K classes and 500K images. For iNaturalist19 and SimpleHierImageNet, we adopt the ID/OOD splits defined in \cite{wallin2025prohoc}, while for iNaturalist21-Aves, we construct a random ID/OOD split (available in the linked code). We create labeled subsets by sampling up to $n$ images per ID class, using all available samples if fewer than $n$ exist. In our experiments, we use 10, 20, and 50 labels per class.

For iNaturalist19 and iNaturalist21-Aves, we append both the unlabeled training set and the test set with 300 randomly selected images from Office31 \cite{saenko2010adapting}, providing out-of-hierarchy OOD data, \ie, OOD samples belonging at the root of the hierarchy. For SimpleHierImageNet, we do not introduce root-level OOD samples, since its top-level nodes correspond to abstract and broad concepts such as \emph{physical object} and \emph{matter}. More dataset details are available in the supplementary material.

\subsection{Training details}

Training of neural networks involves many configurable settings; for most of these, we follow standard configurations from the literature. Our depth-specific classification heads for the ProHOC predictions consist of four layers with a hidden dimension of 512. The backbone is a DinoV2 ViT-L/14 with registers \cite{darcet2023vision}. All models are trained for 400 epochs using stochastic gradient descent with momentum 0.9 and weight decay of 0.001. The labeled batch size is 128, and the unlabeled batch size is set to four times this value. The teacher model is updated as an exponential moving average of the student parameters with momentum 0.999. For pseudo-labeling, we use the threshold $\tau=0.95$.

The hyperparameters we find show sensitivity are the learning rate and dropout. In our experiments, we use a dropout rate of 0.3 throughout. We use a learning rate of 0.01 on iNat19 and iNat21-Aves, and 0.1 for SimpleHierImageNet. A more detailed discussion on the choice of learning rate and dropout is provided in \cref{sec:lr-dropout}.

\subsection{Main results}

To the best of our knowledge, no prior methods have been developed for semi-supervised hierarchical open-set classification. To show the effectiveness of SemiHOC, we compare against several intuitive baselines: 1) \textbf{Supervised only}: a standard pipeline of self-supervised pretraining followed by supervised training of classification heads; 2) \textbf{SSL (Node PLs)}: semi-supervised training from pseudo-labels assigned by thresholding ProHOC predictions per node; and 3) \textbf{SSL (per depth)}: training the depth-specific heads independently using FixMatch-style pseudo-labels per depth. For comparison, we also include \textbf{Sup. (all ID labels)}, which is equal to Supervised only but uses all ID labels, and \textbf{SPL oracle}, which follows SemiHOC without age-gating but with access to correct subtree pseudo-labels for all unlabeled data. All methods are trained on top of frozen DinoV2 features, which have proven to be strong feature representations for our evaluated benchmarks \cite{oquab2023dinov2}. All compared methods use equal hyperparameters within each evaluated dataset. Detailed descriptions of the included baselines are provided in the supplementary material.

Results are shown in \cref{tab:main-results}, evaluated using the balanced mean hierarchical distance (BMHD) \cite{wallin2025prohoc} measured between predictions and ground-truths. Our primary performance metric is BMHD Mix, which averages over ID and OOD performance. We also report the ID and OOD components separately for completeness. Predictions are obtained from the teacher models via ProHOC.

SemiHOC consistently achieves the best performance among methods using limited labels. On iNat19, it even matches the model with full ID supervision when trained with only 20 labels per class, and surpasses it when using 50 labels per class. Notably, the SPL oracle outperforms supervised only with full ID supervision on ID test data, indicating that accurate subtree pseudo-labels for OOD samples provide a valuable learning signal that also benefits ID performance. 

\begin{table}[t]
    \centering
    \caption{Balanced mean hierarchical distance (BMHD) on ID, OOD, and mixed ID+OOD test sets for SemiHOC and baseline methods. The best results under limited labels are \textbf{boldfaced}.} \vspace*{-0.2cm}
    \label{tab:main-results}
    \footnotesize
    \setlength{\tabcolsep}{2pt}
    \begin{tabular}{l >{\color{gray}}c >{\color{gray}}c c >{\color{gray}}c >{\color{gray}}c c >{\color{gray}}c >{\color{gray}}c c}
\toprule
& \multicolumn{3}{c}{10 labels / class} & \multicolumn{3}{c}{20 labels / class} & \multicolumn{3}{c}{50 labels / class} \\ 
\cmidrule(lr){2-4} \cmidrule(lr){5-7} \cmidrule(lr){8-10}
\multicolumn{1}{r}{BMHD $\downarrow$} & ID & OOD & Mix & ID & OOD & Mix & ID & OOD & Mix \\ 
\midrule

& \multicolumn{9}{c}{\textsc{iNaturalist19}} \\ 
\midrule
Sup. (all ID labels) & 0.72 & 0.71 & 0.71 & 0.72 & 0.71 & 0.71 & 0.72 & 0.71 & 0.71 \\
SSL (SPL oracle)              & 0.58 & 0.37 & 0.48 & 0.58 & 0.37 & 0.48 & 0.58 & 0.37 & 0.48 \\ 
\cdashlinelr{1-10}
Supervised only               & 0.82 & 0.84 & 0.83 & 0.71 & 0.81 & 0.76 & 0.71 & 0.73 & 0.72 \\
SSL (Node PLs)                & 0.81 & 0.94 & 0.87 & 0.71 & 0.88 & 0.79 & 0.64 & 0.81 & 0.72 \\
SSL (per depth)    & 0.86 & 1.11 & 0.99 & 0.70 & 1.00 & 0.85 & 0.62 & 0.86 & 0.74 \\
SemiHOC                       & 0.81 & 0.73 & \textbf{0.77} & 0.70 & 0.69 & \textbf{0.69} & 0.64 & 0.62 & \textbf{0.63} \\
\midrule

& \multicolumn{9}{c}{\textsc{iNaturalist21-Aves}} \\ 
\midrule
Sup. (all ID labels) & 0.77 & 0.73 & 0.75 & 0.77 & 0.73 & 0.75 & 0.77 & 0.73 & 0.75 \\
SSL (SPL oracle)              & 0.66 & 0.53 & 0.59 & 0.66 & 0.53 & 0.59 & 0.66 & 0.53 & 0.59 \\ 
\cdashlinelr{1-10}
Supervised only               & 0.89 & 1.00 & 0.94 & 0.79 & 0.95 & 0.87 & 0.74 & 0.85 & 0.79 \\
SSL (Node PLs)                & 0.91 & 0.90 & 0.91 & 0.80 & 0.90 & 0.85 & 0.76 & 0.80 & 0.78 \\
SSL (per depth)    & 0.93 & 1.28 & 1.10 & 0.82 & 1.19 & 1.01 & 0.73 & 1.02 & 0.88 \\ 
SemiHOC                       & 0.90 & 0.82 & \textbf{0.86} & 0.80 & 0.87 & \textbf{0.84} & 0.75 & 0.76 & \textbf{0.76} \\ 
\midrule

& \multicolumn{9}{c}{\textsc{SimpleHierImageNet}} \\ 
\midrule
Sup. (all ID labels) & 0.75 & 1.36 & 1.05 & 0.75 & 1.36 & 1.05 & 0.75 & 1.36 & 1.05 \\
SSL (SPL oracle)              & 0.57 & 0.59 & 0.58 & 0.57 & 0.59 & 0.58 & 0.57 & 0.59 & 0.58 \\ 
\cdashlinelr{1-10}
Supervised only               & 0.82 & 1.88 & 1.35 & 0.72 & 1.73 & 1.23 & 0.69 & 1.65 & 1.17 \\
SSL (Node PLs)                & 0.82 & 1.80 & \textbf{1.31} & 0.72 & 1.66 & 1.19 & 0.68 & 1.58 & 1.13 \\
SSL (per depth)    & 0.93 & 2.36 & 1.65 & 0.86 & 2.07 & 1.47 & 0.67 & 1.70 & 1.19 \\ 
SemiHOC                       & 0.83 & 1.79 & \textbf{1.31} & 0.72 & 1.64 & \textbf{1.18} & 0.68 & 1.55 & \textbf{1.12} \\ 
\bottomrule

\end{tabular}
 \vspace*{-0.5cm}
\end{table}

\subsection{Learning rate and dropout} \label{sec:lr-dropout}

Selecting hyperparameters in settings that involve OOD predictions is not straightforward since we cannot assume access to a test set containing OOD data. Moreover, HND inherently requires balancing performance between ID and OOD data. For example, maximizing ID performance can be achieved trivially by always predicting leaf classes, but this leads to poor OOD predictions.

For SemiHOC, we find that the key hyperparameters for controlling this trade-off are the learning rate and the dropout rate of the student model. Increasing either has a regularizing effect, making the model less confident: this benefits OOD performance but reduces ID performance by decreasing the number of samples predicted as leaf classes.

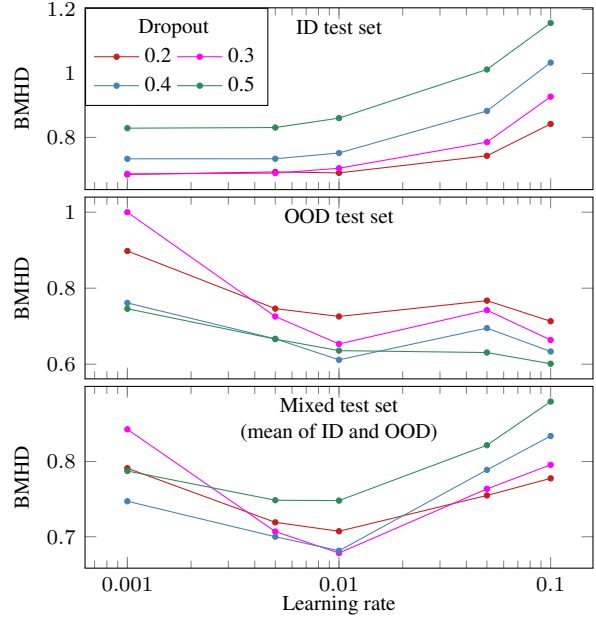
\begin{figure}
    \centering
    \def\datafile{data/semihoc-paramruns_inat19_nlab=20.csv}

\pgfplotsset{
    dropout filter/.style args={#1}{
        x filter/.expression={
            abs(\thisrow{dropout}-#1) > 1e-4 ? NaN : \pgfmathresult
        }
    }
}

\begin{tikzpicture}
\begin{groupplot}[
  group style={group size=1 by 3, vertical sep=1mm},
  width=\columnwidth, height=4cm,
  ylabel=BMHD,
  xmode=log,
  log ticks with fixed point,
  legend cell align=left,
  legend columns=2,
  xlabel style={yshift=5pt},  
  legend style={at={(0.00,1.00)}, anchor=north west, font=\footnotesize},
  every axis plot/.append style={mark=*, mark size=1pt},
  cycle list={{myred},{myfuchsia},{myblue},{mygreen}},
  unbounded coords=discard,
  filter discard warning=false,
  title style={at={(0.5,0.90)},anchor=north}  
]

\nextgroupplot[
    title={ID test set},
    xticklabels={},
    ylabel={BMHD},    
    ]
  \addlegendimage{empty legend}
  \addlegendentry{}

  \addlegendimage{empty legend}
  \addlegendentry{\hspace{-38px}Dropout}

  \foreach \drop in {0.2,0.3,0.4,0.5}{
    \addplot+[]
      table[
        x=lr, y=test_bmhd_id_minhd,
        col sep=comma,
        dropout filter={\drop}
      ] {\datafile};
    \addlegendentryexpanded{\drop}
  }

\nextgroupplot[
    title={OOD test set},
    xticklabels={},    
    ylabel={BMHD},
    ]

  \foreach \drop in {0.2,0.3,0.4,0.5}{
    \addplot+[]
      table[
        x=lr, y=test_bmhd_ood_minhd,
        col sep=comma,
        dropout filter={\drop}
      ] {\datafile};
  }

\nextgroupplot[
    title style = {align = center},
    title={Mixed test set\\(mean of ID and OOD)},
    xlabel=Learning rate,
    ylabel={BMHD},
    ]

  \foreach \drop in {0.2,0.3,0.4,0.5}{
    \addplot+[]
      table[
        x=lr, y=test_bmhd_mix_minhd,
        col sep=comma,
        dropout filter={\drop}
      ] {\datafile};
  }  

\end{groupplot}
\end{tikzpicture} \vspace*{-0.2cm}
    \caption{Increasing the learning or dropout rate has a regularizing effect and improves performance on OOD while decreasing that for ID. The ID and OOD curves share a shoulder point (at around dropout 0.3 and learning rate 0.01), which provides a good trade-off for performance on the mixed ID+OOD setting.} \vspace*{-0.5cm}
    \label{fig:lr-dropout-results}
\end{figure}

\Cref{fig:lr-dropout-results} illustrates this trade-off when varying the learning rate and dropout. Although OOD performance cannot be observed in practice, the curves are roughly symmetric around a region where both ID and OOD performance are strong. This symmetry provides a practical guideline: choose the largest learning rate and dropout value that do not significantly degrade ID performance. For this scenario, iNaturalist with 20 labels per class, a suitable choice is a learning rate of 0.01 and dropout of 0.3. The corresponding results for iNaturalist21-Aves and SimpleHierImagenet are available in the supplementary material.

\subsection{Age-gating} \label{sec:results-age-gating}

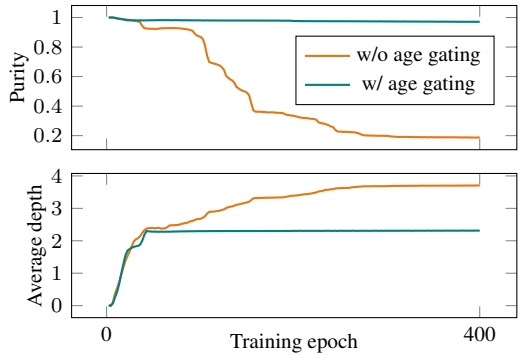
\begin{figure}
    \centering
    \begin{tikzpicture}
\begin{groupplot}[
    group style={
        group size=1 by 2,
        vertical sep=2ex,
    },
    width=0.9\columnwidth,
    height=3.5cm,
    xtick={0,178000},
    xticklabels={0, 400},
    scaled ticks=false,
]

\nextgroupplot[
    ylabel={Purity},
    ylabel style={yshift=-0.5em},
    xticklabels={},    
    legend style={at={(0.95,0.55)}, anchor=east, legend columns=1},
]

\addplot[
    myorange,
    thick
] 
table [
    x=Step,
    y=Value,
    col sep=comma
] {data/run-tensorboard-tag-train_oodset_purity.csv};
\addlegendentry{w/o age gating}

\addplot[
    myteal,
    thick,
] 
table [
    x=Step,
    y=Value,
    col sep=comma
] {data/run-SSLCoarseAgecutoff_entcompprob_thr0.95_dropout0.5-20250707_202227_tensorboard-tag-train_oodset_purity.csv};
\addlegendentry{w/ age gating}

\nextgroupplot[
    ylabel={Average depth},
    xlabel=Training epoch,    
    xlabel style={yshift=1.0em},
    ylabel style={yshift=-0.5em},
]

\addplot[
    myorange,
    thick
] 
table [
    x=Step,
    y=Value,
    col sep=comma
] {data/run-tensorboard-tag-train_oodset_avgdepth.csv};

\addplot[
    myteal,
    thick,
] 
table [
    x=Step,
    y=Value,
    col sep=comma
] {data/run-SSLCoarseAgecutoff_entcompprob_thr0.95_dropout0.5-20250707_202227_tensorboard-tag-train_oodset_avgdepth.csv};

\end{groupplot}
\end{tikzpicture} \vspace*{-0.2cm}
    \caption{Purity and average depth of SPLs assigned to OOD data during training with and without age-gating. Age-gating prevents overconfidence and maintains purity for OOD data throughout training. Results are from iNat19 (10 labels per class).} \vspace*{-0.5cm}
    \label{fig:purity-vs-depth}
\end{figure}

In \cref{sec:age-gating}, we introduced our age-gating strategy for preventing overconfident predictions on OOD data. The key idea is to block samples that are assigned subtree pseudo-labels unexpectedly late. As shown in \cref{fig:purity-vs-depth}, age-gating has a clear effect on the purity of subtree pseudo-labels for OOD data. With age-gating, the purity is maintained throughout training, whereas without it, the model gradually becomes overconfident, leading to a decline in purity.

To further analyze this strategy, we can view age-gating as a binary classifier at each node: correct subtree pseudo-label assignments are treated as positives, and incorrect assignments are negatives. With this perspective, our main objective is to achieve a low false positive rate (FPR), \ie, to minimize the number of incorrect assignments that pass through the gate. However, trivially rejecting all samples would yield zero FPR. We therefore evaluate FPR together with the \emph{coverage}, defined as the fraction of assignments that are not gated, ensuring that a substantial amount of subtree pseudo-labels remains available for training.

\begin{figure}[t]
    \centering
    \pgfplotsset{
    double filter/.style n args={4}{
    x filter/.code={%
      \pgfmathparse{int(\thisrow{#1}) == int(#2) && int(\thisrow{#3}) == int(#4)}
      \ifnum1=\pgfmathresult \pgfmathparse{\thisrow{coverage}}
      \else
        \def\pgfmathresult{inf}%
      \fi
    }
  }    
}

\pgfplotsset{
  covfpr/mark size/.initial=1.8pt,       %
  covfpr/legend mark size/.initial=1.8pt,%
  covfpr/line thickness/.initial=thick,  %
}

\pgfplotsset{
  covfpr/plot base/.style={
    scatter,
    scatter src=explicit symbolic,
    mark size=\pgfkeysvalueof{/pgfplots/covfpr/mark size},
  },
  covfpr/series/.style n args={3}{
    color=#1,
    thick,
    covfpr/plot base,
    double filter={window_size}{#2}{min_peak_size}{#3},
    forget plot,
  },
}

\begin{tikzpicture}

\begin{axis}[
  width=\columnwidth, height=5.2cm,
  xlabel=Coverage, ylabel=FPR,
  grid=both,
  ytick distance=0.2,  
  legend pos=north west,
  xlabel style={yshift=5pt},  
  legend columns=2,
  legend cell align=left,
  legend style={font=\footnotesize, /tikz/column sep=12pt},  
  scatter/classes={
    0.2={mark=square},
    0.1={mark=o},
    0.05={mark=triangle},
    0.03={mark=x},
    0.01={mark=+}
  },  
]

  \addplot+[covfpr/series={myred}{1}{100}]
    table[x=coverage, y=micro_fpr, meta=drop_threshold, col sep=comma]
    {data/inat19cov_vs_fpr_all.csv};

  \addplot+[covfpr/series={myfuchsia}{5}{100}]
    table[x=coverage, y=micro_fpr, meta=drop_threshold, col sep=comma]
    {data/inat19cov_vs_fpr_all.csv};

  \addplot+[covfpr/series={myblue}{10}{100}]
    table[x=coverage, y=micro_fpr, meta=drop_threshold, col sep=comma]
    {data/inat19cov_vs_fpr_all.csv};

  \addplot+[covfpr/series={mygreen}{20}{100}]
    table[x=coverage, y=micro_fpr, meta=drop_threshold, col sep=comma]
    {data/inat19cov_vs_fpr_all.csv};

  \addplot+[covfpr/series={mypurple}{40}{100}]
    table[x=coverage, y=micro_fpr, meta=drop_threshold, col sep=comma]
    {data/inat19cov_vs_fpr_all.csv};

\addlegendimage{empty legend}
\addlegendentry{\hspace{-30pt}bin width $w$}

\addlegendimage{empty legend}
\addlegendentry{\hspace{-30pt}drop threshold $\gamma$}

\addlegendimage{thick, color=myred}
\addlegendentry{$1$}

\addlegendimage{only marks, mark=square, mark size=1.8pt, color=black}
\addlegendentry{$0.20$}

\addlegendimage{thick, color=myfuchsia}
\addlegendentry{$5$}

\addlegendimage{only marks, mark=o, mark size=1.8pt, color=black}
\addlegendentry{$0.10$}

\addlegendimage{thick, color=myblue}
\addlegendentry{$10$}

\addlegendimage{only marks, mark=triangle, mark size=1.8pt, color=black}
\addlegendentry{$0.05$}

\addlegendimage{thick, color=mygreen}
\addlegendentry{$20$}

\addlegendimage{only marks, mark=x, mark size=1.8pt, color=black}
\addlegendentry{$0.03$}

\addlegendimage{thick, color=mypurple}
\addlegendentry{$40$}

\addlegendimage{only marks, mark=+, mark size=1.8pt, color=black}
\addlegendentry{$0.01$}

\end{axis}
\end{tikzpicture} \vspace*{-0.2cm}
    \caption{Analyzing our age-gating performance under varying hyperparameters. The goal is to block incorrect assignments (low FPR) while retaining sufficient data (high coverage).  We obtain results in a favourable trade-off region around 0.8 coverage and 0.1 FPR for many parameter values.} \vspace*{-0.2cm}
    \label{fig:agegating-results}
\end{figure}
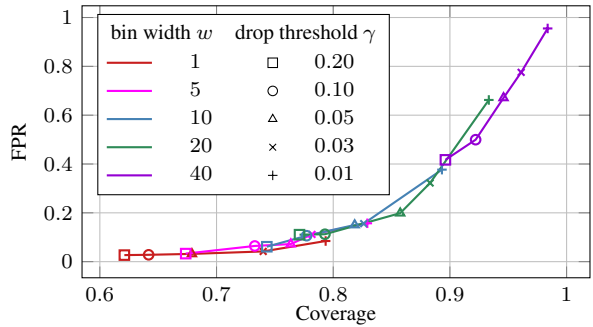

The results of this analysis are summarized in \cref{fig:agegating-results}, where we report FPR and coverage across all assignments for all hierarchy nodes as micro-averages. The results are obtained from a run on iNat19 with 10 labels per class. We vary the hyperparameters of the cutoff-detection in \cref{alg:peak-detection} to study their effect. Increasing the bin width or decreasing the drop threshold allows more samples to pass, pushing both FPR and coverage towards one. Conversely, smaller bin widths and higher drop thresholds result in stricter gating, reducing FPR and coverage. We observe a trade-off region around 0.8 coverage and 0.1 FPR; beyond this point, FPR increases rapidly. Importantly, we note that results in this region are obtained for many parameter pairs, indicating that our age-gating procedure is robust to hyperparameter choices. Unless otherwise stated, we use $w=1$ and $\gamma = 0.01$. The supplementary material contains additional results on age-gating, including analyses of error types and results from other datasets.

\subsection{Ablation}

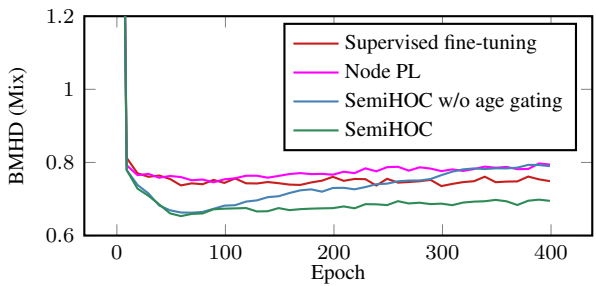
\begin{figure}
    \centering
    \def\datafile{data/ablation_bmhd_mix_minhd.csv}

\begin{tikzpicture}
  \begin{axis}[
    width=\columnwidth, height=4.5cm,
    xlabel={Epoch},
    ylabel={BMHD (Mix)},
    no markers,
    xlabel style={yshift=5pt},
    cycle list={{myred},{myfuchsia},{myblue},{mygreen}},    
    legend cell align=left,
    legend style={at={(0.97,0.97)}, anchor=north east, font=\footnotesize},
    ymin=0.6, ymax=1.2,
    thick,
  ]

    \addplot table [x=Step, y=SupervisedOnly, col sep=comma,] {\datafile};
    \addlegendentry{Supervised fine-tuning}

    \addplot table [x=Step, y=SSLBase, col sep=comma,] {\datafile};
    \addlegendentry{Node PL}

    \addplot table [x=Step, y=SSLCoarse, col sep=comma,] {\datafile};
    \addlegendentry{SemiHOC w/o age gating}

    \addplot table [x=Step, y=SSLCoarseAgecutoff, col sep=comma,] {\datafile};
    \addlegendentry{SemiHOC}

  \end{axis}
\end{tikzpicture} \vspace*{-0.2cm}
    \caption{Performance on iNat19 with 20 labels per class, comparing SemiHOC to baselines with and without age-gating. Subtree pseudo-labels enable SemiHOC to learn efficiently and surpass the baselines, but without age-gating, performance drops over training over training due to overconfident OOD predictions.} \vspace*{-0.5cm}
    \label{fig:ablation}
\end{figure}

To better understand the contributions of subtree pseudo-labels and age-gating, we analyze the performance over training epochs on iNat19 (20 labels per class). We compare SemiHOC with and without age-gating, as well as against baselines. The results in \cref{fig:ablation} show that subtree pseudo-labels provide a strong learning signal, enabling SemiHOC to quickly surpass the baselines. However, without age-gating, performance drops as training progresses due to overconfident predictions on OOD data.

\section{Limitations} \label{sec:limitations}

This work assumes that classes can be organized into a meaningful hierarchy, which does not hold for all classification problems. We further rely on the hierarchy reflecting visually observable concepts; otherwise, OOD samples cannot be expected to be correctly predicted. Finally, our evaluations use the same OOD classes in unlabeled training and test data. An interesting extension for the future is to consider test sets with previously unseen OOD classes.
\section{Conclusion} \label{sec:conclusion}

We introduced semi-supervised hierarchical open-set classification, enabling the use of large unlabeled datasets in HND. Two key challenges were identified: 1) generating reliable pseudo-labels for OOD samples, and 2) preventing overconfident, overly specific predictions of OOD data. Our SemiHOC framework addresses these challenges and achieves strong results across the evaluated benchmarks. We believe there is still room for further work. For example, finding methods for accurately separating OOD data from uncertain ID data would be very valuable in this semi-supervised setting. Moreover, alternative strategies to our age-gating strategy can be explored to address the overconfidence issue. We hope our work can inspire the community to explore new ideas for hierarchical open-set classification.
\section*{Acknowledgement}
\label{sec:acknowledgement}

This work was supported by Saab AB and Wallenberg AI, Autonomous Systems and Software Program (WASP) funded by the Knut and Alice Wallenberg Foundation. The computations were enabled by resources provided by the National Academic Infrastructure for Supercomputing in Sweden (NAISS), partially funded by the Swedish Research Council through grant agreement no. 2022-06725.

{
    \small
    \bibliographystyle{ieeenat_fullname}
    \bibliography{main}

\begin{thebibliography}{37}
\providecommand{\natexlab}[1]{#1}
\providecommand{\url}[1]{\texttt{#1}}
\expandafter\ifx\csname urlstyle\endcsname\relax
  \providecommand{\doi}[1]{doi: #1}\else
  \providecommand{\doi}{doi: \begingroup \urlstyle{rm}\Url}\fi

\bibitem[Bertinetto et~al.(2020)Bertinetto, Mueller, Tertikas, Samangooei, and Lord]{bertinetto2020making}
Luca Bertinetto, Romain Mueller, Konstantinos Tertikas, Sina Samangooei, and Nicholas~A Lord.
\newblock Making better mistakes: Leveraging class hierarchies with deep networks.
\newblock In \emph{Proceedings of the IEEE/CVF Conference on Computer Vision and Pattern Recognition}, 2020.

\bibitem[Cao et~al.(2022)Cao, Brbic, and Leskovec]{cao2022open}
Kaidi Cao, Maria Brbic, and Jure Leskovec.
\newblock Open-world semi-supervised learning.
\newblock In \emph{International Conference on Learning Representations}, 2022.

\bibitem[Darcet et~al.(2024)Darcet, Oquab, Mairal, and Bojanowski]{darcet2023vision}
Timoth{\'e}e Darcet, Maxime Oquab, Julien Mairal, and Piotr Bojanowski.
\newblock Vision transformers need registers.
\newblock In \emph{International Conference on Learning Representations}, 2024.

\bibitem[Fan et~al.(2023)Fan, Kukleva, Dai, and Schiele]{fan2023ssb}
Yue Fan, Anna Kukleva, Dengxin Dai, and Bernt Schiele.
\newblock {SSB}: Simple but strong baseline for boosting performance of open-set semi-supervised learning.
\newblock In \emph{Proceedings of the IEEE/CVF International Conference on Computer Vision}, 2023.

\bibitem[Garg et~al.(2022)Garg, Sani, and Anand]{garg2022learning}
Ashima Garg, Depanshu Sani, and Saket Anand.
\newblock Learning hierarchy aware features for reducing mistake severity.
\newblock In \emph{Proceedings of the European Conference on Computer Vision}, 2022.

\bibitem[Hendrycks and Gimpel(2017)]{hendrycks2016baseline}
Dan Hendrycks and Kevin Gimpel.
\newblock A baseline for detecting misclassified and out-of-distribution examples in neural networks.
\newblock In \emph{International Conference on Learning Representations}, 2017.

\bibitem[{iNaturalist} 2021 competition dataset()]{inaturalist21}
{iNaturalist} 2021 competition dataset.
\newblock {iNaturalist} 2021 competition dataset.
\newblock ~\url{https://github.com/visipedia/inat_comp/tree/master/2021}, 2021.

\bibitem[Jain et~al.(2024)Jain, Karthik, and Gandhi]{jain2024test}
Kanishk Jain, Shyamgopal Karthik, and Vineet Gandhi.
\newblock Test-time amendment with a coarse classifier for fine-grained classification.
\newblock In \emph{Advances in Neural Information Processing Systems}, 2024.

\bibitem[Karthik et~al.(2021)Karthik, Prabhu, Dokania, and Gandhi]{karthik2021no}
Shyamgopal Karthik, Ameya Prabhu, Puneet~K. Dokania, and Vineet Gandhi.
\newblock No cost likelihood manipulation at test time for making better mistakes in deep networks.
\newblock In \emph{International Conference on Learning Representations}, 2021.

\bibitem[Kosmopoulos et~al.(2015)Kosmopoulos, Partalas, Gaussier, Paliouras, and Androutsopoulos]{kosmopoulos2015evaluation}
Aris Kosmopoulos, Ioannis Partalas, Eric Gaussier, Georgios Paliouras, and Ion Androutsopoulos.
\newblock Evaluation measures for hierarchical classification: a unified view and novel approaches.
\newblock \emph{Data Mining and Knowledge Discovery}, 29\penalty0 (3):\penalty0 820--865, 2015.

\bibitem[Lee(2013)]{lee2013pseudo}
Dong-Hyun Lee.
\newblock {Pseudo-Label}: The simple and efficient semi-supervised learning method for deep neural networks.
\newblock In \emph{Proceedings of the ICML Workshop on Challenges in Representation Learning}, 2013.

\bibitem[Lee et~al.(2018)Lee, Lee, Min, Zhang, Shin, and Lee]{lee2018hierarchical}
Kibok Lee, Kimin Lee, Kyle Min, Yuting Zhang, Jinwoo Shin, and Honglak Lee.
\newblock Hierarchical novelty detection for visual object recognition.
\newblock In \emph{Proceedings of the IEEE/CVF Conference on Computer Vision and Pattern Recognition}, 2018.

\bibitem[Li et~al.(2023)Li, Qi, Shi, and Gao]{li2023iomatch}
Zekun Li, Lei Qi, Yinghuan Shi, and Yang Gao.
\newblock {IOMatch}: Simplifying open-set semi-supervised learning with joint inliers and outliers utilization.
\newblock In \emph{Proceedings of the IEEE/CVF International Conference on Computer Vision}, 2023.

\bibitem[Liang et~al.(2018)Liang, Li, and Srikant]{liang2017enhancing}
Shiyu Liang, Yixuan Li, and Rayadurgam Srikant.
\newblock Enhancing the reliability of out-of-distribution image detection in neural networks.
\newblock In \emph{International Conference on Learning Representations}, 2018.

\bibitem[Liang and Davis(2023)]{liang2023inducing}
Tong Liang and Jim Davis.
\newblock Inducing neural collapse to a fixed hierarchy-aware frame for reducing mistake severity.
\newblock In \emph{Proceedings of the IEEE/CVF International Conference on Computer Vision}, 2023.

\bibitem[Linderman et~al.(2023)Linderman, Zhang, Inkawhich, Li, and Chen]{linderman2023fine}
Randolph Linderman, Jingyang Zhang, Nathan Inkawhich, Hai Li, and Yiran Chen.
\newblock Fine-grain inference on out-of-distribution data with hierarchical classification.
\newblock In \emph{Proceedings of the Conference on Lifelong Learning Agents}, 2023.

\bibitem[Liu et~al.(2023)Liu, Wang, Zhang, Fan, Yang, and Shao]{liu2023open}
Jiaming Liu, Yangqiming Wang, Tongze Zhang, Yulu Fan, Qinli Yang, and Junming Shao.
\newblock Open-world semi-supervised novel class discovery.
\newblock In \emph{Proceedings of the International Joint Conference on Artificial Intelligence}, 2023.

\bibitem[Liu et~al.(2020)Liu, Wang, Owens, and Li]{li2020energy}
Weitang Liu, Xiaoyun Wang, John Owens, and Yixuan Li.
\newblock Energy-based out-of-distribution detection.
\newblock In \emph{Advances in Neural Information Processing Systems}, 2020.

\bibitem[Lu et~al.(2023)Lu, Zhu, Zhai, Zheng, and Cao]{lu2023uncertainty}
Fan Lu, Kai Zhu, Wei Zhai, Kecheng Zheng, and Yang Cao.
\newblock Uncertainty-aware optimal transport for semantically coherent out-of-distribution detection.
\newblock In \emph{Proceedings of the IEEE/CVF Conference on Computer Vision and Pattern Recognition}, 2023.

\bibitem[Ma et~al.(2023)Ma, Gao, Zhan, Guo, Zhou, and Wang]{ma2023rethinking}
Qiankun Ma, Jiyao Gao, Bo Zhan, Yunpeng Guo, Jiliu Zhou, and Yan Wang.
\newblock Rethinking safe semi-supervised learning: Transferring the open-set problem to a close-set one.
\newblock In \emph{Proceedings of the IEEE/CVF International Conference on Computer Vision}, 2023.

\bibitem[Miller(1995)]{miller1995wordnet}
George~A Miller.
\newblock Wordnet: a lexical database for english.
\newblock \emph{Communications of the ACM}, 38\penalty0 (11):\penalty0 39--41, 1995.

\bibitem[Oquab et~al.(2024)Oquab, Darcet, Moutakanni, Vo, Szafraniec, Khalidov, Fernandez, HAZIZA, Massa, El-Nouby, Assran, Ballas, Galuba, Howes, Huang, Li, Misra, Rabbat, Sharma, Synnaeve, Xu, Jegou, Mairal, Labatut, Joulin, and Bojanowski]{oquab2023dinov2}
Maxime Oquab, Timoth{\'e}e Darcet, Th{\'e}o Moutakanni, Huy~V. Vo, Marc Szafraniec, Vasil Khalidov, Pierre Fernandez, Daniel HAZIZA, Francisco Massa, Alaaeldin El-Nouby, Mido Assran, Nicolas Ballas, Wojciech Galuba, Russell Howes, Po-Yao Huang, Shang-Wen Li, Ishan Misra, Michael Rabbat, Vasu Sharma, Gabriel Synnaeve, Hu Xu, Herve Jegou, Julien Mairal, Patrick Labatut, Armand Joulin, and Piotr Bojanowski.
\newblock {DINO}v2: Learning robust visual features without supervision.
\newblock \emph{Transactions on Machine Learning Research}, 2024.

\bibitem[Pyakurel and Yu(2024)]{pyakurel2024hierarchical}
Spandan Pyakurel and Qi Yu.
\newblock Hierarchical novelty detection via fine-grained evidence allocation.
\newblock In \emph{Proceedings of the International Conference on Machine Learning}, 2024.

\bibitem[Pyakurel and Yu(2025)]{pyakurellearning}
Spandan Pyakurel and Qi Yu.
\newblock Learning state-based node representations from a class hierarchy for fine-grained open-set detection.
\newblock In \emph{Proceedings of the International Conference on Machine Learning}, 2025.

\bibitem[Rizve et~al.(2022{\natexlab{a}})Rizve, Kardan, Khan, Shahbaz~Khan, and Shah]{rizve2022openldn}
Mamshad~Nayeem Rizve, Navid Kardan, Salman Khan, Fahad Shahbaz~Khan, and Mubarak Shah.
\newblock Openldn: Learning to discover novel classes for open-world semi-supervised learning.
\newblock In \emph{Proceedings of the European Conference on Computer Vision}, 2022{\natexlab{a}}.

\bibitem[Rizve et~al.(2022{\natexlab{b}})Rizve, Kardan, and Shah]{rizve2022towards}
Mamshad~Nayeem Rizve, Navid Kardan, and Mubarak Shah.
\newblock Towards realistic semi-supervised learning.
\newblock In \emph{Proceedings of the European Conference on Computer Vision}, 2022{\natexlab{b}}.

\bibitem[Ruiz and Serrat(2022)]{ruiz2022hierarchical}
Idoia Ruiz and Joan Serrat.
\newblock Hierarchical novelty detection for traffic sign recognition.
\newblock \emph{Sensors}, 22\penalty0 (12):\penalty0 4389, 2022.

\bibitem[Russakovsky et~al.(2015)Russakovsky, Deng, Su, Krause, Satheesh, Ma, Huang, Karpathy, Khosla, Bernstein, et~al.]{russakovsky2015imagenet}
Olga Russakovsky, Jia Deng, Hao Su, Jonathan Krause, Sanjeev Satheesh, Sean Ma, Zhiheng Huang, Andrej Karpathy, Aditya Khosla, Michael Bernstein, et~al.
\newblock Imagenet large scale visual recognition challenge.
\newblock \emph{International Journal of Computer Vision}, 115\penalty0 (3):\penalty0 211--252, 2015.

\bibitem[Saenko et~al.(2010)Saenko, Kulis, Fritz, and Darrell]{saenko2010adapting}
Kate Saenko, Brian Kulis, Mario Fritz, and Trevor Darrell.
\newblock Adapting visual category models to new domains.
\newblock In \emph{Proceedings of the European Conference on Computer Vision}, 2010.

\bibitem[Sohn et~al.(2020)Sohn, Berthelot, Carlini, Zhang, Zhang, Raffel, Cubuk, Kurakin, and Li]{sohn2020fixmatch}
Kihyuk Sohn, David Berthelot, Nicholas Carlini, Zizhao Zhang, Han Zhang, Colin~A Raffel, Ekin~Dogus Cubuk, Alexey Kurakin, and Chun-Liang Li.
\newblock {FixMatch}: Simplifying semi-supervised learning with consistency and confidence.
\newblock In \emph{Advances in Neural Information Processing Systems}, 2020.

\bibitem[Srivastava et~al.(2014)Srivastava, Hinton, Krizhevsky, Sutskever, and Salakhutdinov]{srivastava2014dropout}
Nitish Srivastava, Geoffrey Hinton, Alex Krizhevsky, Ilya Sutskever, and Ruslan Salakhutdinov.
\newblock Dropout: a simple way to prevent neural networks from overfitting.
\newblock \emph{Journal of Machine Learning Research}, 15\penalty0 (1):\penalty0 1929--1958, 2014.

\bibitem[Sun et~al.(2022)Sun, Ming, Zhu, and Li]{sun2022out}
Yiyou Sun, Yifei Ming, Xiaojin Zhu, and Yixuan Li.
\newblock Out-of-distribution detection with deep nearest neighbors.
\newblock In \emph{Proceedings of the International Conference on Machine Learning}, 2022.

\bibitem[Van~Horn et~al.(2018)Van~Horn, Mac~Aodha, Song, Cui, Sun, Shepard, Adam, Perona, and Belongie]{van2018inaturalist}
Grant Van~Horn, Oisin Mac~Aodha, Yang Song, Yin Cui, Chen Sun, Alex Shepard, Hartwig Adam, Pietro Perona, and Serge Belongie.
\newblock The inaturalist species classification and detection dataset.
\newblock In \emph{Proceedings of the IEEE/CVF Conference on Computer Vision and Pattern Recognition}, 2018.

\bibitem[Wallin et~al.(2024)Wallin, Svensson, Kahl, and Hammarstrand]{wallin2024prosub}
Erik Wallin, Lennart Svensson, Fredrik Kahl, and Lars Hammarstrand.
\newblock Prosub: Probabilistic open-set semi-supervised learning with subspace-based out-of-distribution detection.
\newblock In \emph{Proceedings of the European Conference on Computer Vision}, 2024.

\bibitem[Wallin et~al.(2025)Wallin, Kahl, and Hammarstrand]{wallin2025prohoc}
Erik Wallin, Fredrik Kahl, and Lars Hammarstrand.
\newblock Prohoc: Probabilistic hierarchical out-of-distribution classification via multi-depth networks.
\newblock In \emph{Proceedings of the IEEE/CVF Conference on Computer Vision and Pattern Recognition}, 2025.

\bibitem[Yang et~al.(2021)Yang, Wang, Feng, Yan, Zheng, Zhang, and Liu]{yang2021semantically}
Jingkang Yang, Haoqi Wang, Litong Feng, Xiaopeng Yan, Huabin Zheng, Wayne Zhang, and Ziwei Liu.
\newblock Semantically coherent out-of-distribution detection.
\newblock In \emph{Proceedings of the IEEE/CVF International Conference on Computer Vision}, 2021.

\bibitem[Yang et~al.(2024)Yang, Zhou, Li, and Liu]{yang2024generalized}
Jingkang Yang, Kaiyang Zhou, Yixuan Li, and Ziwei Liu.
\newblock Generalized out-of-distribution detection: A survey.
\newblock \emph{International Journal of Computer Vision}, 132\penalty0 (12):\penalty0 5635--5662, 2024.

\end{thebibliography}
}

\clearpage
\setcounter{page}{1}
\maketitlesupplementary

\section{Training algorithms}

To detail the training algorithms for SemiHOC and the baselines, we first describe the depth-specific networks used for ProHOC predictions and the corresponding target distributions for training these.

The depth-specific network at depth $d$, parameterized by $\theta_d$, is responsible for classifying the set of categories at that depth, along with any ID leaf classes that occur at shallower depths. Formally, the classification space at depth $d$ is defined as
\begin{equation}
\begin{aligned}
\mathcal{C}_d = 
&\left\{ c \in \mathcal{C} \mid \text{Depth}(c) = d \right\}
\cup \\
&\left\{ c \in \mathcal{C}_{\text{id}} \mid \text{Depth}(c) < d \right\}
\end{aligned}
\end{equation}
for $d \in \{1,\dots,D\}$ with $D$ being the maximum depth of the hierarchy, $\mathcal{C}$ is the set of all hierarchy nodes, and $\mathcal{C}_\text{id}$ denotes the set of ID leaf classes.

To define the target distributions used for training the depth-specific networks, we introduce a mapping function $S_d(c)$ that returns the ancestors or descendants of $c$ present at depth $d$:
\begin{equation}
    S_d(c) = \left( \text{Anc*}(c) \cup \text{Desc}(c)\right) \cap \mathcal{C}_d,
\end{equation}
where $\text{Anc*}(c)$ denotes the set of ancestors of $c$  (including $c$ itself), and $\text{Desc}(c)$ the set of its descendants. The target distribution $q_c^d(c')$ over $c' \in \mathcal{C}_d$ with $c \in \mathcal{C}$ is then defined as 
\begin{equation}
q_c^d(c') =
\begin{cases}
\dfrac{1}{|S_d(c)|} & \text{if } c' \in S_d(c), \\
0 & \text{otherwise,}
\end{cases}
\end{equation}
where $|\cdot|$ denotes the set cardinality.

When $c$ is an ID class (a leaf), $q_c^d$ reduces to a one-hot distribution over $\mathcal{C}_d$, with the unity element at the ancestor of $c$ at depth $d$. When $c$ is an internal node, $q_c^d$ is one-hot on the ancestors of $c$ for depths $d \leq \text{Depth}(c)$. For deeper levels ($d > \text{Depth}(c)$), the probabilities in $q_c^d$ are spread uniformly over the descendants of $c$ in $\mathcal{C}_d$. The uniform assignment aligns with ProHOC, where the node-local OOD probability is measured through the entropy over the children categories.

SemiHOC is described in \cref{alg:semihoc} and the baselines used for comparisons are described in \cref{alg:supervised,alg:ssl-node,alg:ssl-perdepth}. In these algorithms, \texttt{CE}($\cdot,\cdot$) denotes the cross-entropy loss. Note that \cref{alg:supervised} is equal to the supervised training in \cite{wallin2025prohoc}. Any data augmentations applied during student predictions are absorbed into the student parameters $\theta_d^s$; in our case, this corresponds to applying dropout to the input and intermediate features.

\begin{algorithm}[]
    \scriptsize
    \caption{SemiHOC - training step} \label{alg:semihoc}
    \DontPrintSemicolon
    \SetKwData{DL}{$\{(x^l_i, y_i) \mid i \in \{1,\dots,N\}\}$}
\SetKwData{DU}{$\{(x^{u}_i , g_i)\mid i \in \{1,\dots,M\}\}$}
\SetKwData{LossL}{$\ell^l_d$}
\SetKwData{LossU}{$\ell^{u}_d$}
\SetKwData{Logs}{SPLlog}
\SetKwData{Depth}{depth}
\SetKwData{Sample}{sample}
\SetKwData{Class}{class}
\SetKwData{PredL}{$P^l_{i,d}$}
\SetKwData{PredU}{$P^{u}_{i,d}$}
\SetKwData{HierarchyPreds}{$P_\text{hier}$}
\SetKwData{SubtreeLabels}{$\hat{Y}_{i,d}$}
\SetKwData{Epoch}{epoch}
\SetKwData{None}{none}

\SetKwFunction{Predict}{Predict}
\SetKwFunction{ProHOC}{ProHOC}
\SetKwFunction{ComputeSubtree}{ComputeSubtreePLs}
\SetKwFunction{AddLog}{AddLog}
\SetKwFunction{CE}{CE}

\KwIn{Labeled data: \DL \newline
Unlabeled data with identifiers: \DU \newline
Depth-specific networks: \newline
\hphantom{indent} students $\theta_d^s$, teachers $\theta_d^t$ for $d \in \{1,\dots,D\}$ \newline
Current epoch: $e$ \newline
Age cutoffs: $T_c$ for $c \in \mathcal{C}$ \newline
Subtree pseudo-label log: \Logs
}

\BlankLine
\tcp{Compute the labeled loss}
\For{$d \leftarrow 1$ \KwTo $D$}{
    \LossL $\leftarrow 0$\;
    \LossU $\leftarrow 0$\;
    
    \For{$i \leftarrow 1$ \KwTo $N$}{
        \LossL $\pluseq$ \CE($q_{y_i}^d, p(y|x^l_i;\theta_d^s)$)\;
    }
}

\BlankLine
\tcp{Subtree pseudo-labels for unlabeled data}
\For{$i \leftarrow 1$ \KwTo $M$}{
    \tcp{Teacher predictions}
    $p(y|x^{u}_i) \leftarrow$ \ProHOC($\{p(y|x^{u}_i;\theta^t_d)~|~d \in \{1,\dots,D\}\}$)\; 
    $\mathbf{\hat{y}}_i \leftarrow$ \ComputeSubtree($p(y|x^{u}_i)$) (following \eqref{eq:subtree-pl})\; 
}

\BlankLine
\tcp{Assign pseudo-labels to the log}
\For{$i \leftarrow 1$ \KwTo $M$}{
    \ForEach{$c$ \textbf{where} $\hat{y}_{i,c} = 1$}{
        \If{$(c,g_i)~\text{not in}~\Logs$ }{
            $\Logs(c,g_i) \leftarrow e$\;
        }
    }
    \ForEach{$c$ \textbf{where} $\hat{y}_{i,c} = 0$}{
        \If{$(c,g_i)~\text{in}~\Logs$}{
            Delete $\Logs(c,g_i)$\;
        }
    }
}

\BlankLine
\tcp{Remove any post-cutoff pseudo-labels}
\For{$i \leftarrow 1$ \KwTo $M$}{
    \ForEach{$c$ \textbf{where} $\hat{y}_{i,c}=1$}{
        \If{$\Logs(c,g_i) > T_c$}{
            $\hat{y}_{i,c} \leftarrow 0$\;
        }
    }
}

\BlankLine
\tcp{Add pseudo-labeled data to the unlabeled loss}
\For{$i \leftarrow 1$ \KwTo $M$}{
    \ForEach{$c$ \textbf{where} $\hat{y}_{i,c} = 1$}{
        \ForEach{$d$ \textbf{where} $c \in \mathcal{C}_d$}{
            $\ell_d^u \pluseq \CE(q_c^d, p(y|x^u_i;\theta_d^s))$\;
        }
    }
}

\KwOut{The loss for each depth: \newline
$\ell_d = \ell_d^l/N + \ell_d^u/M$ for $d \in \{1,\dots,D\}$ \newline
Updated \Logs}
\end{algorithm}

\begin{algorithm}[]
    \scriptsize
    \caption{Supervised only - training step} \label{alg:supervised}
    \DontPrintSemicolon
    \SetKwData{DL}{$\{(x^l_i, y_i) \mid i \in \{1,\dots,N\}\}$}
\SetKwData{Loss}{$\ell_d$}
\SetKwData{Depth}{depth}
\SetKwData{Sample}{sample}
\SetKwData{Class}{class}
\SetKwData{PredL}{$P^l_{i,d}$}
\SetKwData{PredU}{$P^{u}_{i,d}$}
\SetKwData{HierarchyPreds}{$P_\text{hier}$}
\SetKwData{SubtreeLabels}{$\hat{Y}_{i,d}$}
\SetKwData{Epoch}{epoch}
\SetKwData{None}{none}

\SetKwFunction{Predict}{Predict}
\SetKwFunction{ProHOC}{ProHOC}
\SetKwFunction{ComputeSubtree}{ComputeSubtreePLs}
\SetKwFunction{AddLog}{AddLog}
\SetKwFunction{CE}{CE}

\KwIn{Labeled data: \DL \newline
Depth-specific networks: $\theta_d$ for $d \in \{1,\dots,D\}$ \newline
}

\BlankLine
\For{$d \leftarrow 1$ \KwTo $D$}{
    \Loss $\leftarrow 0$\;
    
    \For{$i \leftarrow 1$ \KwTo $N$}{
        \Loss $\pluseq$ \CE($q_{y_i}^d, p(y|x^l_i;\theta_d)$)\;
    }
    \Loss $\leftarrow$ $\ell_d/N$
}

\KwOut{The loss for each depth: $\ell_d$ for $d \in \{1,\dots,D\}$}
\end{algorithm}

\begin{algorithm}[]
    \scriptsize
    \caption{SSL (Node PLs) - training step} \label{alg:ssl-node}
    \DontPrintSemicolon
    \SetKwData{DL}{$\{(x^l_i, y_i) \mid i \in \{1,\dots,N\}\}$}
\SetKwData{DU}{$\{x^{u}_i \mid i \in \{1,\dots,M\}\}$}
\SetKwData{LossL}{$\ell^l_d$}
\SetKwData{LossU}{$\ell^{u}_d$}
\SetKwData{Logs}{PLlog}
\SetKwData{Depth}{depth}
\SetKwData{Sample}{sample}
\SetKwData{Class}{class}
\SetKwData{PredL}{$P^l_{i,d}$}
\SetKwData{PredU}{$P^{u}_{i,d}$}
\SetKwData{HierarchyPreds}{$P_\text{hier}$}
\SetKwData{SubtreeLabels}{$\hat{Y}_{i,d}$}
\SetKwData{Epoch}{epoch}
\SetKwData{None}{none}

\SetKwFunction{Predict}{Predict}
\SetKwFunction{ProHOC}{ProHOC}
\SetKwFunction{ComputeSubtree}{ComputeSubtreePLs}
\SetKwFunction{AddLog}{AddLog}
\SetKwFunction{CE}{CE}

\KwIn{Labeled data: \DL \newline
Unlabeled data: \DU \newline
Depth-specific networks: \newline
\hphantom{indent} students $\theta_d^s$, teachers $\theta_d^t$ for $d \in \{1,\dots,D\}$ \newline
Pseudo-label threshold: $\tau$ \newline
}

\BlankLine
\tcp{Compute the labeled loss}
\For{$d \leftarrow 1$ \KwTo $D$}{
    \LossL $\leftarrow 0$\;
    \LossU $\leftarrow 0$\;
    
    \For{$i \leftarrow 1$ \KwTo $N$}{
        \LossL $\pluseq$ \CE($q_{y_i}^d, p(y|x^l_i;\theta_d^s)$)\;
    }    
}

\BlankLine
\tcp{Node pseudo-labels for unlabeled data}
\For{$i \leftarrow 1$ \KwTo $M$}{
    \tcp{Teacher predictions}
    $p(y|x^{u}_i) \leftarrow$ \ProHOC($\{p(y|x^{u}_i;\theta^t_d)~|~d \in \{1,\dots,D\}\}$)\;
    $\hat{y}_i \leftarrow \text{argmax}_y \left[ p(y|x^{u}_i) \right]$\;
}

\BlankLine
\tcp{Add pseudo-labeled data to the unlabeled loss}
\For{$1 \leftarrow 1$ \KwTo $M$}{
    \If{$p(\hat{y}_i | x^{u}_i) > \tau$}{
        \For{$d \leftarrow 1$ \KwTo $D$}{
            \LossU $\pluseq$ \CE($q_{\hat{y}_i}^d, p(y|x^u_i;\theta^s_d)$)
        }
    }
}

\KwOut{The loss for each depth: \newline
$\ell_d = \ell_d^l/N + \ell_d^u/M$ for $d \in \{1,\dots,D\}$}
\end{algorithm}

\begin{algorithm}[]
    \scriptsize
    \caption{SSL (per depth) - training step} \label{alg:ssl-perdepth}
    \DontPrintSemicolon
    \SetKwData{DL}{$\{(x^l_i, y_i) \mid i \in \{1,\dots,N\}\}$}
\SetKwData{DU}{$\{x^{u}_i \mid i \in \{1,\dots,M\}\}$}
\SetKwData{LossL}{$\ell^l_d$}
\SetKwData{LossU}{$\ell^{u}_d$}
\SetKwData{Logs}{PLlog}
\SetKwData{Depth}{depth}
\SetKwData{Sample}{sample}
\SetKwData{Class}{class}
\SetKwData{PredL}{$P^l_{i,d}$}
\SetKwData{PredU}{$P^{u}_{i,d}$}
\SetKwData{HierarchyPreds}{$P_\text{hier}$}
\SetKwData{SubtreeLabels}{$\hat{Y}_{i,d}$}
\SetKwData{Epoch}{epoch}
\SetKwData{None}{none}

\SetKwFunction{Predict}{Predict}
\SetKwFunction{ProHOC}{ProHOC}
\SetKwFunction{ComputeSubtree}{ComputeSubtreePLs}
\SetKwFunction{AddLog}{AddLog}
\SetKwFunction{CE}{CE}

\KwIn{Labeled data: \DL \newline
Unlabeled data: \DU \newline
Depth-specific networks: \newline
\hphantom{indent} students $\theta_d^s$, teachers $\theta_d^t$ for $d \in \{1,\dots,D\}$ \newline
Pseudo-label threshold: $\tau$ \newline
}

\BlankLine
\For{$d \leftarrow 1$ \KwTo $D$}{
    \LossL $\leftarrow 0$\;
    \LossU $\leftarrow 0$\;

    \BlankLine
    \tcp{Labeled data}
    \For{$i \leftarrow 1$ \KwTo $N$}{
        \LossL $\pluseq$ \CE($q_{y_i}^d), p(y|x^l_i;\theta^s_d)$)\;
    }

    \tcp{Unlabeled data}    
    \For{$i \leftarrow 1$ \KwTo $M$}{
        $\hat{y}_i \leftarrow \text{argmax}_y  \left[ p(y|x^u_i;\theta^t_d) \right]$\;
        \If{$p(\hat{y}_i|x^u_i;\theta_d^t) > \tau$}{
            \LossU $\pluseq$ \CE($q_{\hat{y}_i}^d, p(y|x^u_i;\theta^s_d)$)
        }
    }
}

\KwOut{The loss for each depth: \newline
$\ell_d = \ell_d^l/N + \ell_d^u/M$ for $d \in \{1,\dots,D\}$}
\end{algorithm}

\section{Learning rate and dropout for additional datasets}

\begin{figure}[t]
    \centering
    \def\datafile{data/semihoc-paramruns_inat21-birds_nlab=20.csv}

\pgfplotsset{
    dropout filter/.style args={#1}{
        x filter/.expression={
            abs(\thisrow{dropout}-#1) > 1e-4 ? NaN : \pgfmathresult
        }
    }
}

\begin{tikzpicture}
\begin{groupplot}[
  group style={group size=1 by 3, vertical sep=1mm},
  width=\columnwidth, height=4.5cm,
  ylabel=BMHD,
  xmode=log,
  log ticks with fixed point,
  legend cell align=left,
  legend columns=2,
  xlabel style={yshift=5px},
  legend style={at={(0,1.00)}, anchor=north west, font=\footnotesize},
  every axis plot/.append style={mark=*, mark size=1pt},
  cycle list={{myred},{myfuchsia},{myblue},{mygreen}},
  unbounded coords=discard,
  filter discard warning=false,
  title style={at={(0.5,0.90)},anchor=north}  
]

\nextgroupplot[
    title={ID test set},
    xticklabels={},
    ylabel={BMHD},    
    ]

  \addlegendimage{empty legend}
  \addlegendentry{}

  \addlegendimage{empty legend}
  \addlegendentry{\hspace{-38px}Dropout}  

  \foreach \drop in {0.2,0.3,0.4,0.5}{
    \addplot+[]
      table[
        x=lr, y=test_bmhd_id_minhd,
        col sep=comma,
        dropout filter={\drop}
      ] {\datafile};
    \addlegendentryexpanded{\drop}
  }

\nextgroupplot[
    title={OOD test set},
    xticklabels={},    
    ylabel={BMHD},
    ]

  \foreach \drop in {0.2,0.3,0.4,0.5}{
    \addplot+[]
      table[
        x=lr, y=test_bmhd_ood_minhd,
        col sep=comma,
        dropout filter={\drop}
      ] {\datafile};
  }

\nextgroupplot[
    title style = {align = center},
    title={Mixed test set\\(mean of ID and OOD)},
    xlabel=Learning rate,
    ylabel={BMHD},
    ]

  \foreach \drop in {0.2,0.3,0.4,0.5}{
    \addplot+[]
      table[
        x=lr, y=test_bmhd_mix_minhd,
        col sep=comma,
        dropout filter={\drop}
      ] {\datafile};
  }  

\end{groupplot}
\end{tikzpicture} \vspace{-0.3cm}
    \caption{Varying learning rate and dropout for SemiHOC on iNaturalist21-Aves with 20 labels per class.} \vspace{0.5cm}
    \label{fig:lr-dropout-inat21}
    \def\datafile{data/semihoc-paramruns_imagenet_nlab=20.csv}

\pgfplotsset{
    dropout filter/.style args={#1}{
        x filter/.expression={
            abs(\thisrow{dropout}-#1) > 1e-4 || \thisrow{lr} > 0.25 ? NaN : \pgfmathresult
        }
    }
}

\begin{tikzpicture}
\begin{groupplot}[
  group style={group size=1 by 3, vertical sep=1mm},
  width=\columnwidth, height=4.5cm,
  ylabel=BMHD,
  xmode=log,
  log ticks with fixed point,
  legend cell align=left,
  legend columns=2,
  xlabel style={yshift=5px},
  legend style={at={(0,1.00)}, anchor=north west, font=\footnotesize},
  every axis plot/.append style={mark=*, mark size=1pt},
  cycle list={{myred},{myfuchsia},{myblue},{mygreen}},
  unbounded coords=discard,
  filter discard warning=false,
  title style={at={(0.5,0.90)},anchor=north}  
]

\nextgroupplot[
    title={ID test set},
    xticklabels={},
    ylabel={BMHD},    
    ]
  \addlegendimage{empty legend}
  \addlegendentry{}

  \addlegendimage{empty legend}
  \addlegendentry{\hspace{-38px}Dropout}    

  \foreach \drop in {0.1,0.2,0.3,0.4}{
    \addplot+[]
      table[
        x=lr, y=test_bmhd_id_minhd,
        col sep=comma,
        dropout filter={\drop}
      ] {\datafile};
    \addlegendentryexpanded{\drop}
  }

\nextgroupplot[
    title={OOD test set},
    xticklabels={},    
    ylabel={BMHD},
    ]

  \foreach \drop in {0.1,0.2,0.3,0.4}{
    \addplot+[]
      table[
        x=lr, y=test_bmhd_ood_minhd,
        col sep=comma,
        dropout filter={\drop}
      ] {\datafile};
  }

\nextgroupplot[
    title style = {align = center},
    title={Mixed test set\\(mean of ID and OOD)},
    xlabel=Learning rate,
    ylabel={BMHD},
    ]

  \foreach \drop in {0.1,0.2,0.3,0.4}{
    \addplot+[]
      table[
        x=lr, y=test_bmhd_mix_minhd,
        col sep=comma,
        dropout filter={\drop}
      ] {\datafile};
  }  

\end{groupplot}
\end{tikzpicture} \vspace{-0.3cm}
    \caption{Varying learning rate and dropout for SemiHOC on SimpleHierImagenet with 20 labels per class.}
    \label{fig:lr-dropout-imagenet}
\end{figure}

\begin{table*}[t]
    \centering
    \caption{Statistics of our evaluated benchmark datasets.}
    \label{tab:dataset-details}
    \footnotesize
    \begin{tabular}{lrrrrrrrr}
\toprule
Dataset & \# ID train & \# OOD train & \# ID test & \# OOD test & \# Nodes & Depth & \# ID classes & \# OOD classes \\
\midrule
SimpleHierImageNet & 665{,}877 & 100{,}452 & 25{,}900 & 4{,}000 & 582  & 11 & 518 & 80  \\
iNaturalist19      & 156{,}768 & 70{,}738  & 28{,}078 & 12{,}659 & 799  & 6  & 721 & 289 \\
iNaturalist21-Aves & 233{,}062 & 181{,}785 & 8{,}390  & 6{,}470  & 1{,}070 & 4  & 839 & 647 \\
\bottomrule
\end{tabular}
\end{table*}

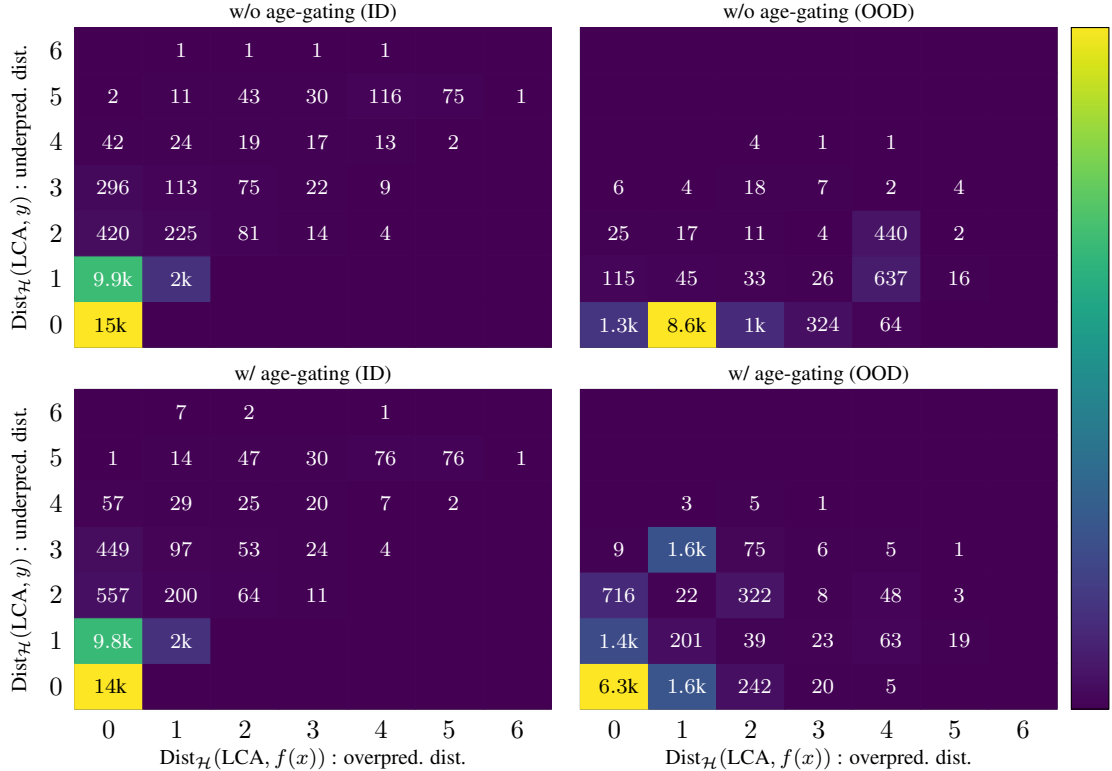
\begin{figure*}[]
    \centering
    \begin{tikzpicture}
\pgfplotsset{compat=1.18}

\begin{groupplot}[
  group style={
    group size=2 by 2,
    vertical sep=1.6em,
    horizontal sep=1.1em,
    xticklabels at=edge bottom,
    yticklabels at=edge left,     
  },
  width=0.95\columnwidth,
  height=0.7\columnwidth,
  ymin=-0.5, ymax=6.5,
  xmin=-0.5, xmax=6.5,
  xtick={0,1,2,3,4,5,6},
  ytick={0,1,2,3,4,5,6},
  hdist base,
  hdist style integers,
  every colorbar/.append style={visualization depends on={}},
  yticklabel style={xshift=3pt},
  xticklabel style={yshift=3pt},
  xlabel style={yshift=3pt},
  title={},
  title style={yshift=-0.27cm},
]

\nextgroupplot[
  ylabel={$\text{Dist}_\mathcal{H}(\text{LCA}, y)$ : underpred. dist.},
  title={w/o age-gating (ID)},
  xlabel={}
]
\addplot [
  matrix plot*,
  mesh/cols=7,
  point meta=explicit,
] table [meta=value, col sep=comma] {data/SSLCoarse_id_hdists.csv};

\nextgroupplot[
  ylabel={},
  title={w/o age-gating (OOD)},
  xlabel={}
]
\addplot [
  matrix plot*,
  mesh/cols=7,
  point meta=explicit,
] table [meta=value, col sep=comma] {data/SSLCoarse_ood_hdists.csv};

\nextgroupplot[
  ylabel={$\text{Dist}_\mathcal{H}(\text{LCA}, y)$ : underpred. dist.},
  title={w/ age-gating (ID)},
  xlabel={$\text{Dist}_\mathcal{H}(\text{LCA}, f(x))$ : overpred. dist.},
]
\addplot [
  matrix plot*,
  mesh/cols=7,
  point meta=explicit,
] table [meta=value, col sep=comma] {data/SSLCoarseAgecutoff_id_hdists.csv};

\nextgroupplot[
  ylabel={},
  title={w/ age-gating (OOD)},
  empty colorbar,
  colorbar style={
    at={(1.03,0)},
    anchor=south west,
    height=2*\pgfkeysvalueof{/pgfplots/parent axis height}+\pgfkeysvalueof{/pgfplots/group/vertical sep}
    },
  xlabel={$\text{Dist}_\mathcal{H}(\text{LCA}, f(x))$ : overpred. dist.}
]
\addplot [
  matrix plot*,
  mesh/cols=7, 
  point meta=explicit,
] table [meta=value, col sep=comma] {data/SSLCoarseAgecutoff_ood_hdists.csv};

\end{groupplot}

\end{tikzpicture}
    \caption{Hierarchical distances between predictions and ground-truths for SemiHOC with and without age-gating. The left panels show ID test data, and the right panels show OOD test data.}
    \label{fig:hdists-inat19}
\end{figure*}

In \cref{sec:lr-dropout} we show results from iNaturalist19, motivating our choice of learning rate and dropout. Here, we show the corresponding results for iNaturalist21-Aves and SimpleHierImageNet. \Cref{fig:lr-dropout-inat21} shows results for iNaturalist21-Aves with 20 labels per class, and \cref{fig:lr-dropout-imagenet} shows results for SimplHierImagenet with 20 labels per class. Following the guideline from \cref{sec:lr-dropout}: choosing the largest learning rate and dropout without significant drops in ID performance, we select a learning rate of 0.01 and dropout of 0.3 for iNaturalist21-Aves, which is the same as for iNaturalist19. For SimpleHierImagenet, we again select dropout 0.3, but ID performance remains stable at higher learning rates, allowing us to choose 0.1.

\section{Dataset details}

\Cref{tab:dataset-details} summarizes the benchmark datasets used in our experiments. For each dataset, we report the number of training and test samples, the number of classes, and hierarchy properties. \emph{\# Nodes} denotes the total number of nodes in the hierarchy, while \emph{Depth} is its maximum depth. \emph{\# ID classes} corresponds to the number of leaf nodes. \emph{\# OOD classes} indicates the number of OOD classes selected from the original dataset. Note that multiple OOD classes may map to the same node in the hierarchy.

\section{Distributions of hierarchical distances}

To further illustrate the overconfidence issue for OOD data in semi-supervised hierarchical open-set classification, \cref{fig:hdists-inat19} shows the distribution of hierarchical distances between predictions and ground-truths for test data at the end of training for SemiHOC, both with and without age-gating. For ID data, the results are similar across the two variants. For OOD data, however, we see big differences. Without age-gating, SemiHOC predicts a majority of OOD samples too deep, with notably many samples predicted as children of the ground-truth. In contrast, with age-gating, it is most common to predict the correct node, and we see a roughly even split between over- and underpredictions. The results are from iNaturalist19 with 20 labels per class.

\section{Generalization of age-gating across datasets}
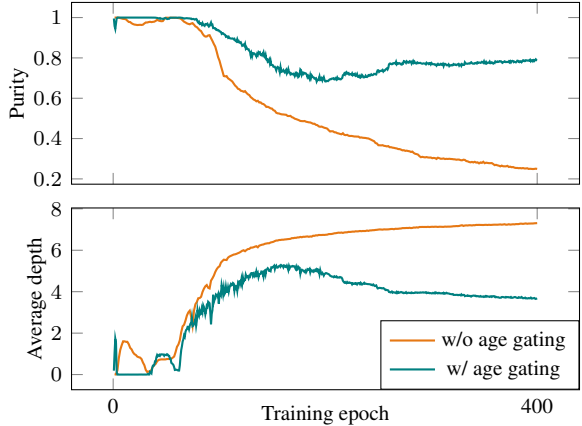
\begin{figure}
    \centering
    \begin{tikzpicture}

\def\agegatingdepth{data/SSLCoarseAgecutoff_imagenet_oodset_avgdepth.csv}
\def\agegatingpurity{data/SSLCoarseAgecutoff_imagenet_oodset_purity.csv}
\def\noagegatingdepth{data/SSLCoarse_imagenet_oodset_avgdepth.csv}
\def\noagegatingpurity{data/SSLCoarse_imagenet_oodset_purity.csv}

\pgfplotstableread[col sep=comma]{\agegatingdepth}\agegatingdepthtable
\pgfplotstableread[col sep=comma]{\noagegatingdepth}\noagegatingdepthtable
\pgfplotstableread[col sep=comma]{\agegatingpurity}\agegatingpuritytable
\pgfplotstableread[col sep=comma]{\noagegatingpurity}\noagegatingpuritytable

\pgfplotstablegetrowsof{\agegatingdepthtable}
\pgfmathtruncatemacro{\lastrowagegating}{\pgfplotsretval-1}

\pgfplotstablegetrowsof{\noagegatingdepthtable}
\pgfmathtruncatemacro{\lastrownoagegating}{\pgfplotsretval-1}

\pgfplotstablegetelem{\lastrowagegating}{Step}\of{\agegatingdepthtable}
\edef\maxstepagegating{\pgfplotsretval}

\pgfplotstablegetelem{\lastrownoagegating}{Step}\of{\noagegatingdepthtable}
\edef\maxstepnoagegating{\pgfplotsretval}

\begin{groupplot}[
    group style={
        group size=1 by 2,
        vertical sep=2ex,
    },
    width=1.0\columnwidth,
    height=4.0cm,
    xtick={0,1},
    xticklabels={0, 400},
    scaled ticks=false,
]

\nextgroupplot[
    ylabel={Purity},
    ylabel style={yshift=-0.5em},
    xticklabels={},
]

\addplot[
    myorange,
    thick
] 
table [
    x=Step,
    x expr={\thisrow{Step}/\maxstepnoagegating},    
    y=Value,
    col sep=comma
] {\noagegatingpuritytable};

\addplot[
    myteal,
    thick,
] 
table [
    x expr={\thisrow{Step}/\maxstepagegating},    
    y=Value,
    col sep=comma
] {\agegatingpuritytable};

\nextgroupplot[
    ylabel={Average depth},
    xlabel=Training epoch,    
    xlabel style={yshift=1.0em},
    ylabel style={yshift=-0.5em},
    legend style={at={(1,0)}, anchor=south east, legend columns=1},    
]

\addplot[
    myorange,
    thick
] 
table [
    y=Value,
    x expr={\thisrow{Step}/\maxstepnoagegating},    
    col sep=comma
] {\noagegatingdepthtable};
\addlegendentry{w/o age gating}

\addplot[
    myteal,
    thick,
] 
table [
    y=Value,
    x expr={\thisrow{Step}/\maxstepagegating},
    col sep=comma
] {\agegatingdepthtable};
\addlegendentry{w/ age gating}

\end{groupplot}
\end{tikzpicture}
    \caption{Purity and average depth of SPLs assigned to OOD data during training with and without age-gating. Results are from SimpleHierImagenet (20 labels per class).}
    \label{fig:purity-vs-depth-imagenet}
\end{figure}

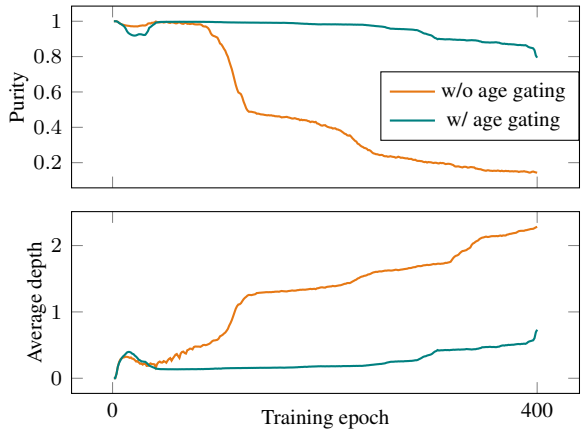
\begin{figure}
    \centering
    \begin{tikzpicture}

\def\agegatingdepth{data/SSLCoarseAgecutoff_inat21_oodset_avgdepth.csv}
\def\agegatingpurity{data/SSLCoarseAgecutoff_inat21_oodset_purity.csv}
\def\noagegatingdepth{data/SSLCoarse_inat21_oodset_avgdepth.csv}
\def\noagegatingpurity{data/SSLCoarse_inat21_oodset_purity.csv}

\pgfplotstableread[col sep=comma]{\agegatingdepth}\agegatingdepthtable
\pgfplotstableread[col sep=comma]{\noagegatingdepth}\noagegatingdepthtable
\pgfplotstableread[col sep=comma]{\agegatingpurity}\agegatingpuritytable
\pgfplotstableread[col sep=comma]{\noagegatingpurity}\noagegatingpuritytable

\pgfplotstablegetrowsof{\agegatingdepthtable}
\pgfmathtruncatemacro{\lastrowagegating}{\pgfplotsretval-1}

\pgfplotstablegetrowsof{\noagegatingdepthtable}
\pgfmathtruncatemacro{\lastrownoagegating}{\pgfplotsretval-1}

\pgfplotstablegetelem{\lastrowagegating}{Step}\of{\agegatingdepthtable}
\edef\maxstepagegating{\pgfplotsretval}

\pgfplotstablegetelem{\lastrownoagegating}{Step}\of{\noagegatingdepthtable}
\edef\maxstepnoagegating{\pgfplotsretval}

\begin{groupplot}[
    group style={
        group size=1 by 2,
        vertical sep=2ex,
    },
    width=1.0\columnwidth,
    height=4.0cm,
    xtick={0,1},
    xticklabels={0, 400},
    scaled ticks=false,
]

\nextgroupplot[
    ylabel={Purity},
    ylabel style={yshift=-0.5em},
    xticklabels={},
    legend style={at={(1,0.45)}, anchor=east, legend columns=1},     
]

\addplot[
    myorange,
    thick
] 
table [
    x=Step,
    x expr={\thisrow{Step}/\maxstepnoagegating},    
    y=Value,
    col sep=comma
] {\noagegatingpuritytable};
\addlegendentry{w/o age gating}

\addplot[
    myteal,
    thick,
] 
table [
    x expr={\thisrow{Step}/\maxstepagegating},    
    y=Value,
    col sep=comma
] {\agegatingpuritytable};
\addlegendentry{w/ age gating}

\nextgroupplot[
    ylabel={Average depth},
    xlabel=Training epoch,    
    xlabel style={yshift=1.0em},
    ylabel style={yshift=-0.5em},   
]

\addplot[
    myorange,
    thick
] 
table [
    y=Value,
    x expr={\thisrow{Step}/\maxstepnoagegating},    
    col sep=comma
] {\noagegatingdepthtable};

\addplot[
    myteal,
    thick,
] 
table [
    y=Value,
    x expr={\thisrow{Step}/\maxstepagegating},
    col sep=comma
] {\agegatingdepthtable};

\end{groupplot}
\end{tikzpicture}
    \caption{Purity and average depth of SPLs assigned to OOD data during training with and without age-gating. Results are from iNaturalist21-Aves (20 labels per class)}
    \label{fig:purity-vs-depth-inat21}
\end{figure}

In \cref{sec:results-age-gating}, we analyzed the effect of age-gating on the purity and depth of subtree pseudo-labels assigned to OOD data in iNaturalist19. Here, we provide the corresponding results for SimpleHierImagenet and iNaturalist21-Aves to demonstrate that the benefits of age-gating generalize across datasets. The results are shown in \cref{fig:purity-vs-depth-imagenet,fig:purity-vs-depth-inat21}. For both datasets, age-gating helps maintain throughout training. On SimpleHierImagenet, a challenging benchmark, the age-gated purity is lower than for the iNaturalist datasets, but the improvement over the no-age-gating baseline remains substantial.

\end{document}